%% file: main.tex
\documentclass{article} %
\usepackage{iclr2025_conference,times}

\input{math_commands.tex}

\usepackage{hyperref}
\usepackage{url}
\usepackage{enumitem}
\usepackage{natbib}
\usepackage{booktabs}       %
\usepackage{amsfonts}       %
\usepackage{graphicx}
\usepackage{adjustbox}
\usepackage{subfigure}
\usepackage{xspace}
\usepackage{placeins}
\usepackage{afterpage}
\usepackage{multirow}
\usepackage{colortbl}
\usepackage{xcolor}

\newcommand{\unet}{UNet\xspace}

\iclrfinalcopy

\title{\centering Slot-Guided Adaptation of Pre-trained Diffusion Models for Object-Centric Learning and Compositional Generation}

\author{
~~~~~~~~~~~~~~~~~~~~~~~~~~~~~~~~Adil Kaan Akan$^1$ ~~~~~~~~~~~~~~~~~~~~~~~~~~~~~~~~~~~~~~~Yucel Yemez$^{1,2}$ \\
~~$^1$Department of Computer Engineering, Koc University~~$^2$KUIS AI Center, Koc University
}

\begin{document}

\maketitle
\begin{abstract}
\vspace{-0.1cm}

We present SlotAdapt, an object-centric learning method that combines slot attention with pretrained diffusion models by introducing adapters for slot-based conditioning. Our method preserves the generative power of pretrained diffusion models, while avoiding their text-centric conditioning bias. We also incorporate an additional guidance loss into our architecture  to align cross-attention from adapter layers with slot attention. This enhances the alignment of our model with the objects in the input image without using external supervision. Experimental results show that our method outperforms state-of-the-art techniques in object discovery and image generation tasks across multiple datasets, including those with real images. Furthermore, we demonstrate through experiments that our method performs remarkably well on complex real-world images for compositional generation, in contrast to other slot-based generative methods in the literature. The project page can be found at \href{https://kaanakan.github.io/SlotAdapt/}{https://kaanakan.github.io/SlotAdapt/}.

\end{abstract} \vspace{-0.5cm}
\input{sections/intro}
\input{sections/rw}
\input{sections/method}

\input{sections/exp}

\input{sections/conclusion}

\boldparagraph{Acknowledgements}We thank Aykut Erdem and Erkut Erdem for their insightful discussions.

\newpage
\bibliography{iclr2025_conference}
\bibliographystyle{iclr2025_conference}

\newpage
\appendix
\section{Appendix}
In this appendix, we provide more details and results about our work. It is organized into three parts:

\begin{itemize}
    \item Training and Architectural Details (\secref{sec:train_details}):
We explain how we trained our model SlotAdapt and describe its structure in depth.
    \item Additional Experiments and Evaluations (\secref{sec:additional_exps}): We share additional experimental and evaluation results
    \item More Examples (\secref{sec:qual}):
We show additional visual examples of SlotAdapt's performance in different settings.
    \item Comparison with GLASS and SPOT (\secref{sec:glass}):
We compare SlotAdapt with GLASS and SPOT, similar studies, to show how our work fits into current research.
    
\end{itemize}

\input{supp/training_details}
\afterpage{\FloatBarrier}
~
\input{supp/rebuttal_exps}

\afterpage{\FloatBarrier}
\newpage
\input{supp/qual_examples}
\afterpage{\FloatBarrier}
\newpage
\input{supp/glass_comp}

\end{document}

%% file: math_commands.tex
\usepackage{amsmath,amsfonts,bm}

\def\Figref#1{Figure~\ref{#1}}

\def\secref#1{section~\ref{#1}}

\def\eqref#1{equation~\ref{#1}}

\def\1{\bm{1}}

\def\rvepsilon{{\mathbf{\epsilon}}}

\DeclareMathAlphabet{\mathsfit}{\encodingdefault}{\sfdefault}{m}{sl}
\SetMathAlphabet{\mathsfit}{bold}{\encodingdefault}{\sfdefault}{bx}{n}

\newcommand{\boldparagraph}[1]{\vspace{0.2cm}\noindent{\bf #1:} }

\newcommand{\xdownarrow}[1]{%
  {\left\downarrow\vbox to #1{}\right.\kern-\nulldelimiterspace}
}

\newcommand{\xuparrow}[1]{%
  {\left\uparrow\vbox to #1{}\right.\kern-\nulldelimiterspace}
}

\newcommand{\eqnref}[1]{Eq.~\ref{#1}}
\newcommand{\tabref}[1]{Table~\ref{#1}}

\newcommand{\bx}{\mathbf{x}}

\newcommand{\bS}{\mathbf{S}}

\newcommand{\bU}{\mathbf{U}}

\newcommand{\bA}{\mathbf{A}}

\newcommand{\bff}{\mathbf{f}}

\newcommand{\br}{\mathbf{r}}

\newcommand{\btheta}{\boldsymbol{\theta}}

\newcommand{\bepsilon}{\boldsymbol{\epsilon}}

%% file: sections/intro.tex
\section{Introduction}
\label{sec:intro} \vspace{-0.35cm}
The real world is inherently structured with distinct, composable parts and objects that can be combined in various ways; this compositional characteristic is essential for cognitive functions like reasoning, understanding causality, and ability to generalize beyond training data \citep{lake2017building,bottou2014machine, scholkopf2021toward, bahdanau2018systematic, fodor88}. While language clearly reflects this modularity through sentences made up of distinct words and tokens, the compositional structure is less obvious in visual data. Object-centric learning (OCL) offers a promising approach to uncover this latent structure by grouping related features into coherent object representations without supervision \citep{kahneman1992reviewing, greff2020binding}. By decomposing complex scenes into separate objects and their interactions, OCL mimics how humans interpret their environment \citep{spelke2007core}, potentially improving the robustness and interpretability of AI systems \citep{lake2017building,scholkopf2021toward}. This approach shifts from traditional pixel-based feature extraction to a more meaningful segmentation of visual data, which is key for better generalization and supporting high-level reasoning tasks.

Recent advances in OCL have shown the potential to incorporate powerful generative models, such as transformers and diffusion models, into the OCL framework as image decoders. Notably, models such as Latent Slot Diffusion (LSD)~\citep{jiang2023lsd} and SlotDiffusion~\citep{wu2023slotdiffusion} have considerably improved performance in object discovery and visual generation tasks in real-world settings by employing slot-conditioned diffusion models. A concurrent work, GLASS ~\citep{singh2024guidedsa}, has made further progress in handling complex natural scenes by using generated captions as  external guiding signals. 

There exist two main approaches of integrating diffusion models into slot-based OCL methods in order to deal with real-world images. The first one, as seen in GLASS and LSD, directly uses a pretrained stable diffusion model as decoder. While this approach attempts to fully harness the generative power of pretrained diffusion models, it relies on cross-attention layers to achieve slot conditioning, which were however optimized to receive text input. In contrast, the second approach, as in SlotDiffusion, trains the diffusion model from scratch on the target dataset, thereby eliminating any such text-conditioning related biases. However, this limits the generation capacity of the slot-conditioned model, making it less effective in handling complex real-world images.

Enabling slot-based methods to fully exploit the generative capabilities of pretrained diffusion models while avoiding biases due to text-trained conditioning is a major challenge in object-centric learning, which hinders the effective handling of complex real-world images, particularly in compositional image generation tasks \citep{jung2024composition}. In this work, we propose a novel approach that addresses this challenge by combining slot attention \citep{locatello2020object} with adaptive conditioning in diffusion models~\citep{mou2024t2i, rombach2022ldm, sohl2015dm, ho2020ddpm}. Our method, that we refer to as SlotAdapt, introduces the use of adapter layers~\citep{mou2024t2i} for slot conditioning, which eliminates text-trained conditioning bias in pretrained diffusion models and allows the slots to diverge from representations in the text embedding space while retaining the generative power of the pretrained diffusion model.  

Another challenge for OCL methods is dealing with the part-whole hierarchy problem – the difficulty of deciding whether to segment an object into its parts or as a whole~\citep{hinton1979some}, especially in real-world settings. The GLASS model ~\citep{singh2024guidedsa} mitigates this problem with the aid of external supervision and at the cost of increased complexity, whereas \cite{jung2024composition} introduce compositional representation learning; however the applicability of their method on real-world datasets remains limited. In order to address this challenge, we propose using cross-attention masks from adapter layers as pseudo-ground truth to guide slot attention maps. This self-supervisory signal enhances the alignment between learned slots and actual objects without external supervision, facilitating the learning of meaningful object representations. In fact, in our architecture, the guidance is mutual, as the cross-attention masks in the diffusion model are also guided and refined jointly with the slot attention masks, improving the image generation process.

In summary, we have the following main contributions: 1) We introduce adapters for slot-based conditioning to combine slot attention with pretrained diffusion models, 2) We propose an additional guidance loss to align cross-attention from adapter layers with slot attention without using external supervision, and 3) We present compositional image generation results on complex real-world images.

Through extensive experiments on various datasets, we demonstrate that our method, SlotAdapt, outperforms previous approaches in object discovery and compositional image generation tasks~\citep{jiang2023lsd, wu2023slotdiffusion, seitzer2023dinosaur}, especially on complex real-world image datasets. The performance of SlotAdapt in segmentation task is similar with the concurrent work~\citep{singh2024guidedsa}, yet we achieve this with significantly reduced computational requirements and without any external supervision. Note that GLASS does not address the compositional generation task. To the best of our knowledge, our work is the first to present successful results for compositional generation on COCO \citep{lin2014coco}, a large complex real-world image dataset with 80 different object classes.

%% file: sections/rw.tex
\vspace{-0.2cm}
\section{Related Work} \vspace{-0.35cm}
\subsection{Unsupervised Object-Centric Learning} \vspace{-0.25cm}
Unsupervised object-centric learning aims to decompose visual scenes into meaningful object representations without need for annotation. Early methods used iterative inference and CNN decoders to reconstruct scenes from object-specific feature vectors (\emph{slots}) \citep{eslami2016air, burgess2019monet, greff2019iodine, lin2019space, jiang2019scalor, lin2020gswm}. Attend-Infer-Repeat (AIR) \citep{eslami2016air} and Sequential AIR (SQAIR) \citep{kosiorek2018sqair} reconstructed objects in canonical poses using patch-based decoders. Slot attention \citep{locatello2020object} and SAVi \citep{kipf2021savi} employed the spatial broadcast decoder \citep{watters2019sbdecoder} to predict images and segmentation masks from slots, combined via alpha compositing. While effective on synthetic datasets like CLEVR \citep{johnson2017clevr} and 3D Shapes \citep{burgess2018_3dshapes}, these methods struggle with complex, real-world images. Transformer-based decoders, such as SLATE \citep{singh2021slate} and STEVE \citep{singh2022steve}, pre-train a discrete VAE (dVAE) to tokenize images and use slot-conditioned transformers for autoregressive reconstruction. Despite improvements, they face challenges in the high-quality reconstruction of complex real images \citep{singh2021slate, wu2023slotformer}. Methods like DINOSAUR \citep{seitzer2023dinosaur} bypass reconstruction by using self-supervised learning to discover objects but lack image generation capabilities. To improve slot-based autoencoders, \citet{Kakogeorgiou2024SPOT} propose self-training and patch-order permutation strategies, enhancing segmentation in complex real-world images. A relatively recent research direction in OCL is to employ diffusion models as slot decoders, enhancing scene decomposition and visual fidelity in real-world scenarios ~\citep{jiang2023lsd, wu2023slotdiffusion}. Several approaches have tackled compositional scene generation through different mechanisms: combining global and symbolic latent variables \citep{jiang2020ref1}, using hierarchical VAEs with slot attention \citep{wang2023ref2}, and leveraging hierarchical discrete representations with autoregressive transformers \citep{wu2024ref3}. While these works focus on synthetic datasets, our work can achieve compositional generation in real-world scenarios.

\subsection{Diffusion Models} \vspace{-0.15cm}
Diffusion models \citep{sohl2015dm, ho2020ddpm} have shown remarkable versatility in tasks such as class-conditioned generation, text-to-image synthesis, and image editing \citep{dhariwal2021adm, ho2021cfdm, meng2021sdedit, ho2022cascadedm,saharia2022srdm, ramesh2022dalle2, nichol2022glide, saharia2022imagen}. Latent Diffusion Models (LDMs) \citep{rombach2022ldm} address the computational complexity of diffusion models by operating in a lower-dimensional latent space through the use of a pre-trained autoencoder. This approach significantly reduces computational load while maintaining generation quality. LDMs also introduce a flexible conditioning mechanism through cross-attention layers, enabling integration of various conditioning information, such as text embeddings. Recent advances such as T2I adapters \citep{mou2024t2i, biner2024sonicdiffusion} further enhance the adaptability of LDMs. By introducing additional adapter cross-attention layers, these methods allow fine-tuning for new conditional tasks while keeping the base LDM frozen, reducing computational costs and data requirements for model adaptation.

\vspace{-0.35cm}
\subsection{Object-Centric Learning Methods with Diffusion Models} \vspace{-0.25cm}
Recently, several works have investigated the use of diffusion models in object-centric learning, such as Latent Slot Diffusion (LSD) \citep{jiang2023lsd}, SlotDiffusion \citep{wu2023slotdiffusion}, \citep{jung2024composition} and GLASS \citep{singh2024guidedsa}. All these methods employ LDMs \citep{rombach2022ldm} as slot decoders. Their approaches are similar in that they all use slot attention \citep{locatello2020object} to extract slot representations from input image and then condition the diffusion model with these slots through cross-attention modules. These methods primarily differ in whether they use pretrained diffusion models, fine-tune them, or train them from scratch. LSD and GLASS use pretrained diffusion models to be able to deal with complex real-world images, assuming an inherent alignment between slot representations and the text-trained cross-attention within the diffusion model, which hinders their performance, especially on image generation tasks. SlotDiffusion, on the other hand, opts to retrain the diffusion model on the target dataset, thereby avoiding biases due to text-trained conditioning, though at the cost of increased computational complexity and, more importantly, at the loss of the generative power of pretrained diffusion models.

GLASS, which is actually a concurrent work  \citep{singh2024guidedsa}, attempts to mitigate the issues due to text-trained conditioning in pretrained diffusion models by utilizing cross-attention masks as pseudo-ground truth to guide slot attention, in a manner similar to our approach. However, it necessitates an external image captioner or class labels to do this and requires two forward passes: one for generating the training images and pseudo masks, and another for slot attention and image reconstruction. Despite this, GLASS still struggles, particularly in differentiating object instances of the same class, and  does not present any compositional image generation results. In contrast, SlotAdapt incorporates additional cross-attention layers into the pretrained diffusion model as adapters, enabling the slots to focus primarily on object semantics, rather than being constrained within a text-centric embedding space.

\cite{jung2024composition} have recently introduced a novel framework for compositional learning, which incorporates an additional compositional path into their architecture alongside the slot-conditioned diffusion model. Their approach processes two images simultaneously, composing objects from these images into a single output to apply a diffusion-based generative prior loss. This facilitates the learning of compositional information; however, the applicability of the resulting model on complex real-world images is limited; more specifically, they present results on BDD100K \citep{BDD}, an autonomous driving dataset with limited context and relatively small number of object classes.

\vspace{-0.4cm}

%% file: sections/method.tex
\begin{figure}[!ht]
\centering
\includegraphics[width=1.\textwidth]{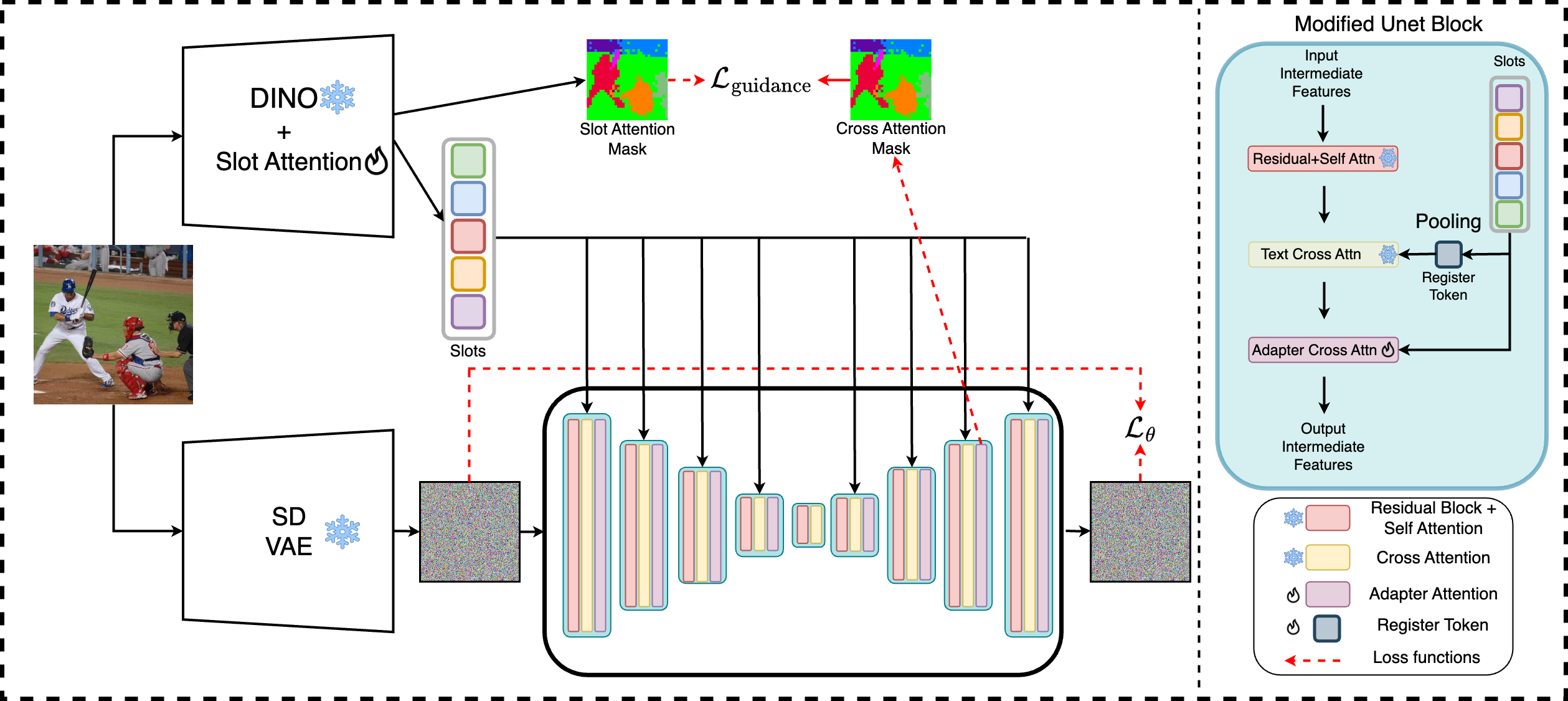}
\vskip -0.1in
\caption{\textbf{SlotAdapt Architecture} 
We extract object-centric information from the input image using a visual backbone, which combines DINO and slot attention. Stable Diffusion VAE is used to encode the image into latent space and then noise is added to the latent. Diffusion process is conditioned on the generated slots as well as the register token which is generated by (mean) pooling the slots. We use the original cross attention layers of diffusion model to condition on the register token, and additional adapter attentions to condition on the learned slots. The overall objective is to predict the noise added to the image. Additionally, we introduce a guidance loss between the slot attention masks and adapter cross attention masks, which encourages the similarity between them. The guidance is only applied in the third upsampling block, while slot conditioning is applied throughout all downsampling and upsampling blocks.
}
\vspace{-0.6cm}
\label{fig:method}
\end{figure}
\section{Slot-Based Object-Centric Learning}
\label{sec:method_sec} 
\vspace{-0.35cm}
\subsection{Background}
\label{sec:background} \vspace{-0.4cm}
\boldparagraph{Slot Attention} Slot Attention \citep{locatello2020object} provides a robust framework for segmenting input data into discrete, interpretable components. It operates by dynamically allocating a set of learnable vectors, termed ``slots", to represent distinct entities within the input data. These slots are often initialized  randomly and then iteratively updated via an attention mechanism, enabling the model to bind each slot to different parts of the input corresponding to individual objects or object-like features. The iterative update rule for each slot is given by
\vspace{-0.85cm}
\begin{center}
\label{eq:slot}
\begin{align}
\bU^{(m)} &= \text{Attention}\left(q(\bS^{(m)}), k(\bff), v(\bff)\right) \\
\bS^{(m+1)} &= \text{GRU}(\bS^{(m)}, \bU^{(m)}) \nonumber
\end{align}
\end{center}
\vspace{-0.25cm} where $q$, $k$, $v$ are learnable linear functions, respectively corresponding to the query, key and value in attention computation; $\bff$ denotes image features; $\bS$ is the set of slots; $\bU$ represents the update generated by the attention operation; and $m$ is iteration number for the GRU (Gated Recurrent Unit). The initial slots, $\bS^{(0)}$, can be initialized from a Gaussian distribution or using initial object locations~\citep{kipf2021savi, elsayed2022savi++}. Attention is computed over the slot dimension, causing slots to compete for pixels. As a result, each slot attempts to bind to a distinct object in each iteration.
The effectiveness of Slot Attention lies in its capacity to disentangle and encode complex scenes into structured representations without supervision. This is achieved through the ability of the mechanism to assign representational capacity where needed. 

\boldparagraph{Diffusion Models} 
Diffusion models \citep{sohl2015dm, ho2020ddpm} are a powerful class of generative models that simulate the gradual addition of noise to data and then learn to reverse this process. Once trained, they can generate high-quality samples from pure Gaussian noise, and have shown remarkable success in various domains, particularly in computer vision for challenging tasks such as image generation and editing.

The generative process involves a series of reverse diffusion steps from a noise distribution $p(\mathbf{z})$ to the data distribution $p(\mathbf{x})$, defined by conditional probabilities $p(\mathbf{x}_{t-1} | \mathbf{x}_t)$. The training objective minimizes the difference between the actual and generated distributions, typically using a loss function $\mathcal{L}$ that penalizes the expected error of noise prediction at each time step $t$:
\begin{equation}
\label{eq:denoise}
\mathcal{L({\btheta})} = \mathbb{E}_{\mathbf{x} \sim p(\mathbf{x}), {\bepsilon}_t\sim \mathcal{N}(0,1), t \sim \mathcal{U}(1, T)}\left[\Vert {\bepsilon}_t - {\epsilon}_{{\btheta}}(\mathbf{x}_t, t, y) \Vert_{2}^2\right],
\end{equation}
where ${\epsilon}_{\btheta}$ is the noise prediction of the neural network with parameters ${\btheta}$, and $y$ is a conditioning signal for the model such as class or text. When dealing with images, the common practice is to perform training and sampling in a low-dimensional latent space, which is achieved by first transforming the input via a pretrained variational encoder and then to use the standard U-Net architecture as the denoising neural network ${\bepsilon}_{\theta}$ \citep{rombach2022ldm}.

Previous OCL methods which employ diffusion models as slot decoders replace the condition $y$ in \eqnref{eq:denoise} with slot representations generated by Slot Attention \citep{locatello2020object} and use reconstruction loss as a learning signal.\vspace{-0.25cm}
\subsection{SlotAdapt}
\label{sec:slot_adapt} \vspace{-0.35cm}
\boldparagraph{Object-centric Visual Encoding}
We start with an input image $\bx \in \mathcal{R}^{H \times W \times 3}$ and transform it through a visual backbone combined with slot attention in a manner similar to recent works \citep{jiang2023lsd, wu2023slotdiffusion}. The visual backbone serves as a feature extractor, reducing the image size to create a set of visual feature vectors $\bff \in \mathcal{R}^{h \cdot w \times d}$ by flattening. This reduction makes computations more efficient and condenses key features of the image into a more compact representation. The slot attention mechanism works on these vectors, using a competitive process to generate $N$ slots, represented as $\bS \in \mathcal{R}^{N \times d}$. Each slot contains information about a separate object or entity in the scene. The resulting slot representations are then used to condition the diffusion-based decoder as explained in the following section (see also Fig.~\ref{fig:method}).

\boldparagraph{Slot Conditional Decoding}
We employ a pretrained Stable Diffusion \unet model as a decoder conditioned on the extracted slot representations. This \unet is primarily composed of residual, self-attention, and cross-attention layers stacked together.
While self-attention layers capture spatial information, cross-attention layers model semantic relationships between text embeddings and intermediate features. 
Inspired by recent works \citep{mou2024t2i, ye2023ipadapter}, we extend this architecture by introducing an adapter layer after each existing cross-attention layer in all downsampling and upsampling blocks of the \unet. These adapter layers are dedicated to conditioning the model on the extracted slots and are essentially cross-attention layers which function similarly to their text-based counterparts in the \unet, with a few minor differences (see Appendix for implementation details).  

Our adapter-based conditioning strategy is substantially different from the conventional use of text-trained cross-attention for slot conditioning. Our rationale is that by including these additional cross-attention layers, we enable the slots to focus primarily on object semantics, rather than being constrained within a text-centric embedding space. This is particularly crucial, given that the cross-attention layers in pretrained diffusion modules are typically optimized for text embeddings and hence expect textual input.

We also introduce an extra token that leverages the unused text-conditioning modules of the pretrained \unet to better capture context. This extra token is computed by (mean) pooling either the generated slots or the image features from the visual backbone, and then fed as input into the text cross-attention layers of the \unet model. The main motivation here is to create a ‘register’ token, similar to the idea presented by \cite{darcet2023register}. This register acts as a storage for overall scene information within the diffusion model. By giving this task to a dedicated token, we allow the individual slots to concentrate more on specific objects instead of diluting their focus with background or contextual details.

We freeze the pretrained diffusion model and train only the adapter layers and the slot attention component of our architecture. The training process minimizes a reconstruction objective, formulated as a noise prediction problem. Following \citep{rombach2022ldm}, at each training iteration (assuming a mini-batch of size 1), time step $t$, latent image $\mathbf{x}$ and noise ${\bepsilon}_t$ are first sampled from their respective distributions. The noisy image $\mathbf{x}_t$ is then calculated via forward diffusion and fed into the denoising \unet  $\rvepsilon_{\btheta}$. The \unet predicts the noise ${\bepsilon}_t$, conditioned on the extracted slots $\bS$ and the register token $\br$, and is updated based on the gradient of the following loss function (see also Eq.~\ref{eq:denoise}):
\begin{equation}
\centering
\mathcal{L}_t(\btheta) = \Vert \bepsilon_t - \rvepsilon_{\btheta}(\bx_t, t, \bS, \br) \Vert_{2}^{2}
\end{equation}
\boldparagraph{Attention Guidance} The attention mask generated by slot attention serves as an affinity measure between image features and slot vectors, effectively segmenting objects in the image under the assumption that the slots capture object representations. We enhance this framework by leveraging the cross-attention mask extracted from the adapter layers in the diffusion model as a self-supervisory signal to guide slot attention, and/or vice versa. These dual attention masks, one from slot attention and the other from the diffusion model, encode similar semantics regarding the relationship between slot representations and image features. Normally, there are as many adapter cross-attention masks as there are \unet blocks, but we focus only on the one from the third upsampling block (the second-to-last) since this layer is very close to the output and empirically provides an attention mask most aligned with the objects in the image.  We denote the slot attention mask by $\bA_{\text{\tiny SA}}$ and the diffusion attention mask with $\bA_{\text{\tiny DM}}$. In both cases, the attention masks are computed through the dot product between queries and keys, and normalized over the query dimension. In slot attention, slots act as queries and image features as keys, whereas in the attention mechanism of the diffusion model, this relationship is inverted. Formally we can write
\begin{equation}
\label{eq:attention_mask}
    \bA_{\text{\tiny SA}} = \mathrm{Softmax} \left(\frac{k_{\text{\tiny SA}}(\bff_{\text{\tiny SA}}) q_{\text{\tiny SA}}(\bS)^\top}{\sqrt{D}}\right) \quad \quad \bA_{\text{\tiny DM}} = \mathrm{Softmax} \left(\frac{k_{\text{DM}}(\bS) q_{\text{\tiny DM}}(\bff_{\text{\tiny DM}})^\top}{\sqrt{D}}\right)
\end{equation}
where $q_{\text{\tiny SA}}$, $k_{\text{\tiny SA}}$, $q_{\text{\tiny DM}}$, and $k_{\text{\tiny DM}}$ are learnable linear functions, and $\bff_{\text{\tiny SA}}$ and $\bff_{\text{\tiny DM}}$ represent image features. The cross-attention layer in the adapter employs a multi-head structure, resulting in multiple attention masks. We average these masks over the head dimension and use the result as the diffusion attention mask $\bA_{\text{\tiny DM}}$.

The slot attention mask, $\bA_{\text{\tiny SA}} \in \mathcal{R}^{(h \cdot w) \times N}$, encodes how each pixel in the image features relates to the slot representations. Ideally, this attention mask should converge to an instance segmentation mask, with each slot representing a distinct object.
Conversely, the diffusion attention mask, $\bA_{\text{\tiny DM}} \in \mathcal{R}^{N \times (h \cdot w)}$, shows the inverse relationship. In the optimal scenario, we expect this attention map to converge to the transpose of the instance segmentation mask, enabling the diffusion model to generate the input image accurately. We formulate a guidance loss to enforce the alignment of these dual attention masks by
\begin{equation}
\mathcal{L}_{\text{guidance}} = \text{BCE}(\bA_{\text{\tiny SA}}, \bA_{\text{\tiny DM}}^\top),
\end{equation}
where BCE is the binary cross-entropy loss.  The overall training objective combines this guidance loss with the primary loss function, weighted by a hyperparameter $\lambda$:
\begin{equation}
\mathcal{L} = \mathcal{L}_{\btheta} + \lambda \mathcal{L}_{\text{guidance}}
\end{equation}%
 There are different design choices for implementing the guidance loss: 1) guiding only the slot attention mask $\bA_{\text{\tiny SA}}$ with $\bA_{\text{\tiny DM}}$ (stopping gradient for diffusion model), 2) guiding only the diffusion attention mask $\bA_{\text{\tiny DM}}$ with $\bA_{\text{\tiny SA}}$ (stopping gradient for slot attention),  and 3) joint guidance, that is, guiding both $\bA_{\text{\tiny SA}}$ and $\bA_{\text{\tiny DM}}$ simultaneously (no gradient stopping). So in the first two options, one attention mask serves as a pseudo ground-truth for the other, whereas in the last option, the two masks are jointly enforced for alignment.
 
 Another alternative we have considered is guiding the attention masks through multiplication. In this scenario, there is no explicit auxiliary loss: each of the cross-attention masks from multiple attention heads in the adapter layer is simply multiplied by $\bA_{\text{\tiny SA}}$ using the Hadamard product. Since each cross-attention mask ideally represents a part of the object bound by the corresponding slot, the multiplication confines the adapter attention to the region defined by the slot attention mask.   

 All the guidance alternatives described above aim, in one way or another, to enhance the alignment of the attention matrices which represent the same semantics in the architecture, thereby improving their alignment with the objects in the input image. We compare these different alternatives for attention mask guidance in the experiments section.

%% file: sections/exp.tex
\vspace{-0.3cm}

\section{Experiments}
\label{sec:exp}
\vspace{-0.5cm}
\boldparagraph{Datasets}
Our evaluation framework covers both synthetic and real-world datasets, aligning with recent works in the field \citep{jiang2023lsd, wu2023slotdiffusion}. We assess our method SlotAdapt on the synthetic MOVi-E dataset \citep{greff2022kubric} and two widely-recognized real-world datasets: VOC \citep{everingham2010pascalvoc} and COCO \citep{lin2014coco}.
While our primary focus is on leveraging pretrained diffusion models for real-world scenarios, we include MOVi-E in our evaluation due to its complexity, featuring scenes with up to 23 objects. This dataset serves as a challenging benchmark for object-centric learning in controlled environments.
Both real-world datasets, VOC and COCO, have emerged as popular benchmarks for object discovery tasks. They present significant challenges due to their multi-object nature and the large number of foreground classes they contain—20 and 80, respectively. These datasets have been instrumental in recent evaluations of various object-centric learning methods on real-world images \citep{jiang2023lsd, wu2023slotdiffusion, seitzer2023dinosaur}.

\boldparagraph{Implementation Details} Following the previous works \citep{jiang2023lsd, wu2023slotdiffusion}, we use a convolutional backbone for MOVi-E and DINOv2~\citep{oquab2023dinov2} with ViT-B and a patch size of 14 as the encoder model for VOC and COCO. To serve as our decoder, we incorporate a pretrained Stable Diffusion (SD) model, v1.5 for MOVi-E; and COCO and v2.1 for VOC~\citep{rombach2022ldm}, augmented with an additional cross-attention layer as adapter. For MOVi-E, we jointly optimize the image backbone, slot attention mechanism, and adapter layers. For VOC and COCO, we focus our training exclusively on the slot attention and adapter layers. We maintain consistency with previous work \citep{jiang2023lsd, wu2023slotdiffusion} in terms of dataset selection and preparation. Our models are trained for approximately 150K to 250K iterations. While this training duration is shorter compared to some previous works, we demonstrate that our approach achieves competitive performance.

\input{tabs/archiecture_ablations}

\vspace{-0.3cm}\boldparagraph{Baselines}We compare SlotAdapt with unsupervised state-of-the-art object-centric methods. On the MOVi-E dataset, we compare with SLATE~\citep{singh2021slate}, SLATE$^+$, where SLATE's low capacity dVAE is replaced by a pre-trained VQGAN model~\citep{esser2021vqgan} and Latent Slot Diffusion~(LSD)~\citep{jiang2023lsd}. On real-world datasets, we compare with Slot attention~(SA)~\citep{locatello2020object}, SLATE~\citep{singh2021slate}, DINOSAUR~\citep{seitzer2023dinosaur}, LSD~\citep{jiang2023lsd} and SlotDiffusion~\citep{wu2023slotdiffusion}. In all our experiments on real-world datasets, we use a frozen DINOv2~\citep{caron2021dino,oquab2023dinov2} as the visual encoder for all models.

\vspace{-0.3cm}\boldparagraph{Metrics} Following previous work~\citep{locatello2020object, jiang2023lsd}, we employ a set of standard metrics to assess the performance of our method on unsupervised object segmentation. These include the foreground adjusted rand index (FG-ARI), mean intersection over union (mIoU), and mean best overlap (mBO). Our evaluation is performed on the slot attention masks, $\bA_\text{slot}$, computed as in Equation \ref{eq:attention_mask}. We use two different versions of the mIoU and mBO metrics: one computed over instance-level masks, and the other over class-level masks. Note that instance-level metrics account for whether objects of the same class in an image are differentiated as separate instances in the resulting segmentation; hence, they are more informative on the representational and generative capabilities of an object-centric learning method. On the other hand, the FG-ARI, a metric designed primarily for object discovery task, does not account for object masks larger than the ground-truth, which may be problematic, especially when assessing the generative capability.

\vspace{-0.3cm}
\begin{figure*}[hb]
\centering
\includegraphics[width=.8\textwidth]{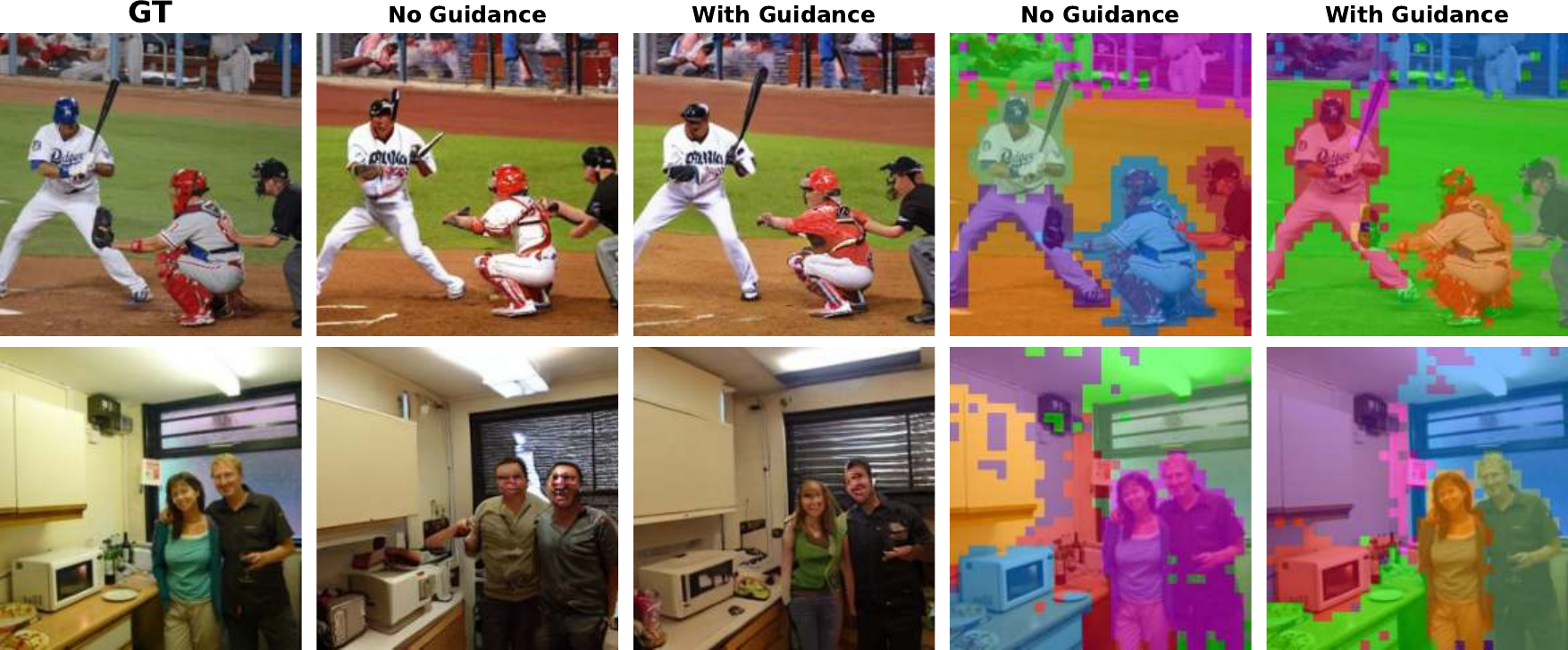}
\vskip -0.1in
\caption{\textbf{Qualitative comparison: with vs. without guidance.} We visualize generated images and predicted segments on COCO dataset.}
\vspace{-0.5cm}
\label{fig:guidance_compare}
\end{figure*}

\subsection{Ablation Studies}\vspace{-0.2cm}
\label{sec:ablation}
To assess the impact of our contributions, we conduct a series of experiments. All the experiments in this ablation study are conducted with register token (slot pooling), joint guidance of attention masks and conditioning through all downsampling and upsampling blocks, unless stated otherwise.

We first investigate the optimal block combination in the \unet architecture for conditioning the diffusion model on slots, where the options we consider are all downsampling blocks, all upsampling blocks, mid-block and their certain combinations. The results obtained on MOVi-E dataset are presented in the left part of \tabref{tab:combined_ablation}. We observe that conditioning on either the upsampling blocks alone or both the downsampling and upsampling blocks yield superior performance, likely due to their proximity to the input and output in terms of structure and resolution. 

Next, we evaluate the effect of incorporating a register token \citep{darcet2023register} into the textual cross-attention layer in the diffusion model. The results obtained on MOVi-E are given in the right part of \tabref{tab:combined_ablation}), where the options are no register token, slot pooling and feature pooling. We observe that inclusion of register token, with slot or feature pooling, yields consistent improvements in all metrics, including segmentation accuracy and performance in downstream tasks.

Lastly, we examine the effectiveness of our guidance strategies on COCO dataset in \tabref{tab:guidance-ablation}. We observe that joint guidance yields the best performance on all metrics except the FG-ARI score, where the improvements are substantial compared to the no-guidance case. For FG-ARI metric, multiplication guidance gives the best performance, which is a sufficient metric for object discovery but not necessarily for generative tasks. Moreover, in Fig.~\ref{fig:guidance_compare}, we visually demonstrate the impact of the joint guidance strategy on the generated segmentation masks. We observe that the inclusion of guidance yields a significant improvement in segmentation quality, mitigating the part-whole hierarchy problem and producing segmentation masks better aligned with the objects in the scene, rather than with partial or fragmented objects. In turn, the generated (reconstructed) images are more faithful to the original input images, as also observed in Fig.~\ref{fig:guidance_compare}.  \vspace{-0.3cm}
\input{tabs/guidance_ablation}
\subsection{Evaluation Results}%
All the evaluation experiments for SlotAdapt are conducted with register token, joint guidance and conditioning on all downsampling and upsampling blocks, unless stated otherwise.\vspace{-0.1cm}

\vspace{-0.4cm}
\begin{figure*}[hb]
\centering
\includegraphics[width=1.\textwidth]{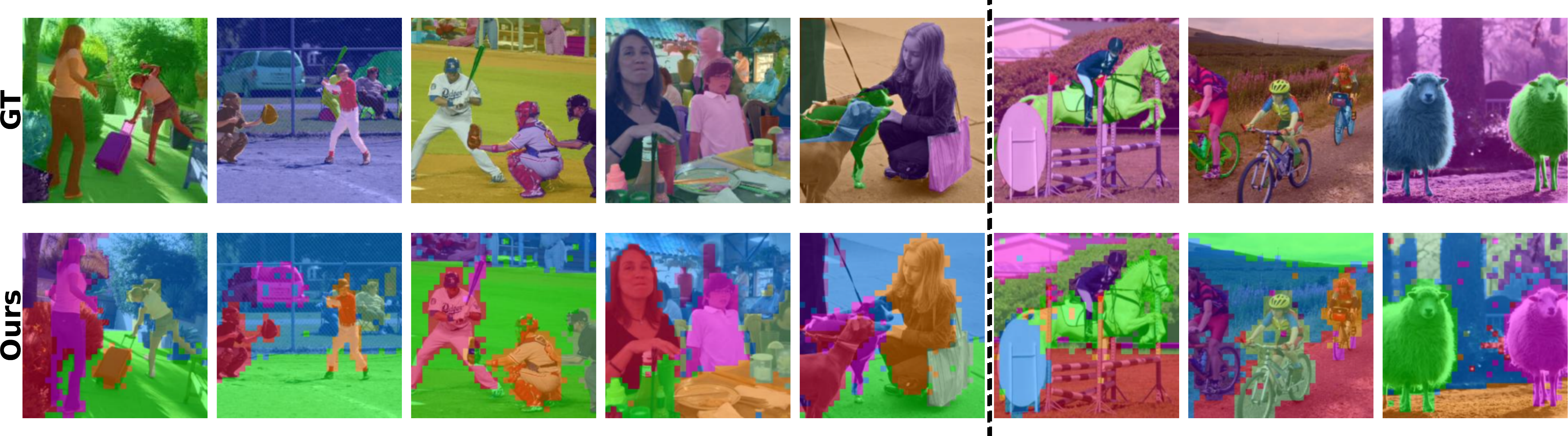}
\vskip -0.2in
\caption{\textbf{Unsupervised Object Segmentation.}We show visualizations of segments on COCO~(left) and VOC~(right). SlotAdapt accurately binds distinct instances belonging to the same class.}

\label{fig:seg}
\end{figure*}

\vspace{-0.2cm}\boldparagraph{Synthetic Dataset} We evaluate the object discovery and segmentation performance of our method on MOVi-E in comparison to the baseline methods in \tabref{tab:movie}, which is presented in Appendix, based on the FG-ARI, instance-level mBO and instance-level mIoU metrics. We observe significant improvements in almost all metrics, particularly with an approximate 10\% enhancement in object discovery and segmentation accuracy.

\vspace{-0.2cm}\boldparagraph{Real-World Dataset} We evaluate the object discovery and segmentation performance of our method on the COCO and VOC datasets in \tabref{table:real-world-seg-quan}, using the FG-ARI, instance-level mBO and class-level mBO metrics, in comparison to baseline methods including StableLSD~\citep{jiang2023lsd}, SlotDiffusion~\citep{wu2023slotdiffusion}, and DINOSAUR~\citep{seitzer2023dinosaur}. 

We observe that SlotAdapt, when used with guidance, outperforms all the baselines for almost all metrics, including the highly competitive DINOSAUR method. The only exception is the class-level mBO score on VOC, where SlotAdapt performs worse than SlotDiffusion. Notably, the improvement in mBO scores on COCO is particularly significant, about 10\% better than the next best baseline. The impact of guidance is also substantial on COCO, which is a more challenging and much larger dataset with complex multi-object scenes and varying object sizes, when compared to VOC.

\begin{figure*}[hb!]
\centering
\includegraphics[width=1.\textwidth]{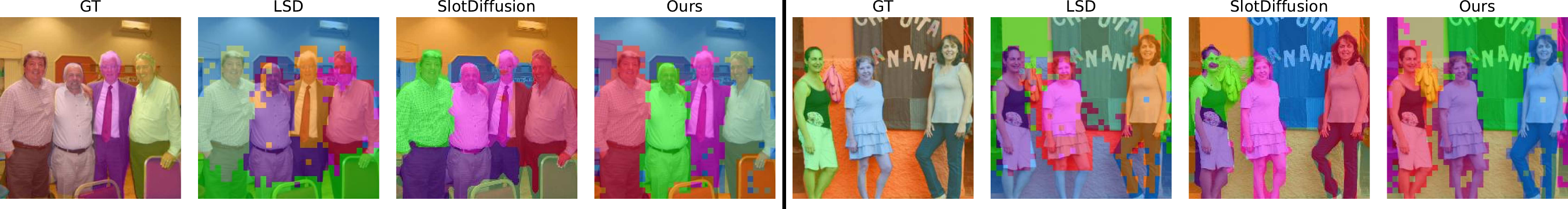}
\vskip -0.2in
\caption{\textbf{Qualitative comparisons with other methods on COCO.} We visualize predicted segments of SlotAdapt in comparison to LSD and SlotDiffusion. SlotAdapt can more effectively differentiate between object instances of the same class compared to other methods.}
\vspace{-0.5cm}
\label{fig:seg_compare}
\end{figure*}

In Fig.\ref{fig:seg_compare}, we visually compare the segmentation results of LSD, SlotDiffusion and SlotAdapt on COCO. We observe that SlotAdapt successfully differentiates individual object instances as  evidenced by its superior instance-level mBO score, whereas LSD and Slot Diffusion struggle with this challenge. Moreover, SlotAdapt  produces more complete segmentations of objects without dividing them into parts, which is reflected by its higher FG-ARI score. These results highlight the robustness and versatility of SlotAdapt in handling the complexities of real-world data.

\begin{figure}[!h]
\centering
\includegraphics[width=1.\textwidth]{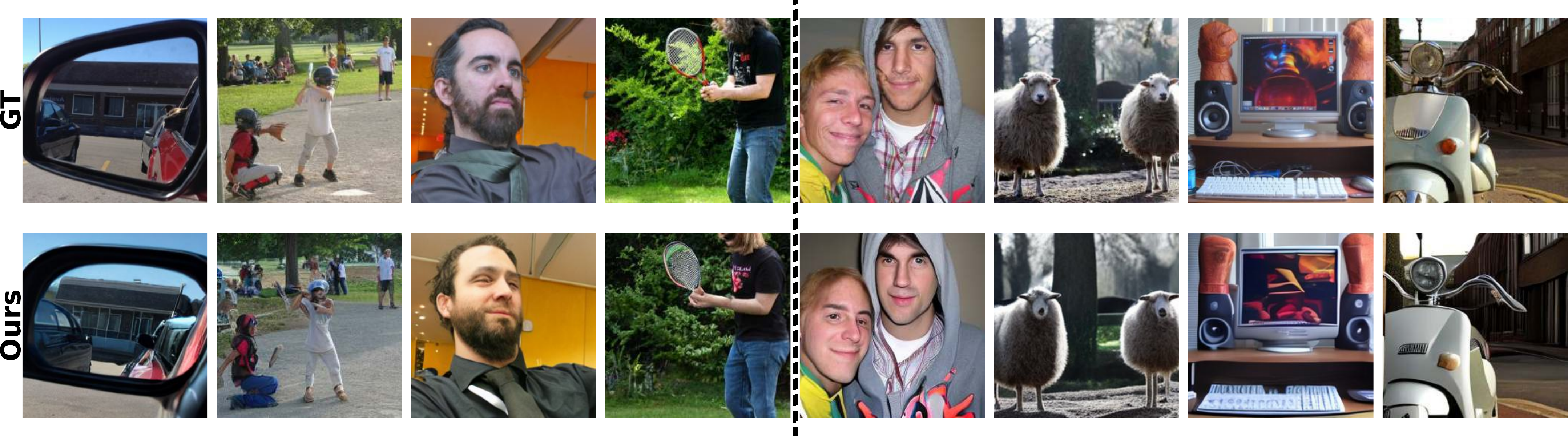}
\vskip -0.1in
\caption{\textbf{Generation Results.}We show sample images reconstructed by SlotAdapt on COCO~(left) and VOC~(right). SlotAdapt generates reconstructions highly faithful to the original input images.}
\vspace{-0.5cm}
\label{fig:gen}
\end{figure}

\begin{figure}[ht]
\centering
\includegraphics[width=1.\textwidth]{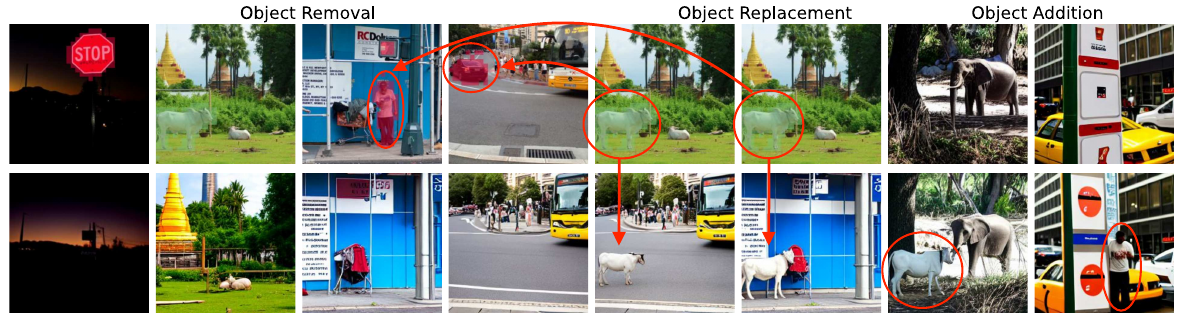}
\vskip -0.15in
\caption{\textbf{Compositional Editing.} We demonstrate object removal, replacement and addition edits on COCO images by using slots. Removing highlighted slots (top row) yields realistic and successful generations (first 4 examples). Replacing highlighted objects in the 3rd and 4th images with the cow object from the 5th and 6th images results in highly accurate edits, yet with small changes in the original images. Finally, adding the cow  (5th image) and the person (3rd image) slots to the last two images, respectively, generates meaningful examples of complex scenes.
\vspace{-0.4cm}}
\label{fig:comp_gen}
\end{figure}

\vspace{-0.2cm}\boldparagraph{Generation and Compositional Editing} We first demonstrate the ability of our model to generate realistic images on COCO in Fig.~\ref{fig:gen}. We see that, when conditioned on slots, our model can reconstruct the input image with high quality, realism and notable fidelity. Regarding compositional generation and editing capabilities, Figure \ref{fig:comp_gen} shows a series of image edits by modifying input slots, including object replacement, removal and addition. We observe that the editing operations are highly successful and seamless with only slight yet realistic changes to the image background, while all maintaining high quality. To assess our model's capabilities, we conducted extensive experiments evaluating both its reconstruction fidelity and compositional generation performance. The comprehensive results are presented in Table~\ref{tab:model_eval}. For additional qualitative and quantitative results, we refer readers to Section~\ref{sec:qual} in the appendix.

\vspace{-0.35cm}

\input{tabs/generation_eval}

\input{tabs/coco_voc}

%% file: tabs/archiecture_ablations.tex
\begin{table}[!h]
\centering
\vspace{-0.5cm}
\caption{\textbf{Architectural ablations on MOVi-E.} We examine the effects of architectural choices on segmentation and representation performance. We present block combinations on the left and register token choice on the right. Up, Down and Mid refer to all upsampling blocks, all downsampling blocks and middle block in the diffusion model.}
\label{tab:combined_ablation}
\resizebox{\textwidth}{!}{%
\begin{tabular}{@{}lccccc|ccc@{}}
\toprule
& \multicolumn{5}{c}{Conditioning Blocks}& \multicolumn{3}{c}{Register Token} \\
\cmidrule(lr){2-6} \cmidrule(lr){7-9}
& Up+Down+Mid & Only Up & Only Down & Up+Mid & Up+Down & No Token & Slot Pooling & Feature Pooling \\
\midrule
\multicolumn{9}{c}{Segmentation (\%)} \\
\midrule
FG-ARI (\textuparrow) & 56.89 & 57.38 & \textbf{57.93} & 57.39 & 56.45 & 54.38 & 56.27 & \textbf{57.18} \\
mBO (\textuparrow) & 39.59 & 43.05 & 40.20 & 39.96 & \textbf{43.38} & 40.07 & 41.65 & \textbf{43.98} \\
mIoU (\textuparrow) & 37.75 & 41.53 & 38.77 & 38.83 & \textbf{41.86} & 40.07 & \textbf{40.10} & 39.83 \\
\midrule
\multicolumn{9}{c}{Representation} \\
\midrule
Category (\textuparrow) & 43.92 & 43.82 & \textbf{45.88} & 41.54 & 43.91 & 42.42 & \textbf{43.54} & 42.63 \\
Position (\textdownarrow) & 1.92 & 1.82 & \textbf{1.61} & 1.92 & 1.72 & 1.89 & \textbf{1.75} & 1.78 \\
3D B-Box (\textdownarrow) & 3.94 & 3.78 & \textbf{3.48} & 3.95 & 3.75 & 3.83 & \textbf{3.77} & 3.78 \\
\bottomrule
\end{tabular}%
} \vspace{-0.2cm}
\end{table}

%% file: tabs/guidance_ablation.tex
\begin{table}[t]
    \caption{
    \textbf{Evaluation of guidance strategies.} We present the segmentation performance on COCO for different guidance strategies. Joint guidance gives the best scores and significantly improves over no guidance option.
    \label{tab:guidance-ablation}
    }
    \centering
        \centering
        \small
        \setlength{\tabcolsep}{5pt}
        \begin{tabular}{lccccc}
        \toprule
        \textbf{} & FG-ARI & mBO$^i$ & mBO$^c$ & mIoU$^i$ & mIoU$^c$ \\
        \midrule
        No guidance & 42.3 & 31.5 & 34.8 & 31.7 & 38.5 \\
        \midrule
        Slot Guidance & 41.2 & 33.4 & 36.9 & 33.1 & 37.9 \\
        DM Guidance & 42.0 & 31.2 & 34.6 & 32.0 & 38.4 \\
        Joint Guidance & 41.4 & \textbf{35.1} & \textbf{39.2} & \textbf{36.1} & \textbf{41.4} \\
        Multiplication Guidance & \textbf{43.3} & 31.9 & 35.3 & 31.7 & 36.4  \\
        \bottomrule
        \end{tabular}
        \vspace{-0.6cm}
\end{table}

%% file: tabs/generation_eval.tex
\begin{table}[t]
    \centering
    \small
    \caption{\textbf{Model Evaluation} Comparison of FID and KID scores for reconstruction (left) and compositional generation (right) across different methods.}
    \label{tab:model_eval}
    \setlength{\tabcolsep}{5pt}
    \begin{minipage}[t]{0.45\textwidth}
        \begin{tabular}{lc|c}
        \toprule
        Method & FID & KID$\times$1000 \\
        \midrule
        LSD & 35.537 & 19.086 \\
        SlotDiffusion & 19.448 & 5.852 \\
        Ours & \textbf{10.857} & \textbf{0.388} \\
        \bottomrule
        \end{tabular}
    \end{minipage}%
    \hspace{0.5cm} %
    \begin{minipage}[t]{0.45\textwidth}
        \begin{tabular}{lc|c}
        \toprule
        Method & FID & KID$\times$1000 \\
        \midrule
        LSD & 167.232 & 103.482 \\
        SlotDiffusion & 64.213 & 57.309 \\
        Ours & \textbf{40.568} & \textbf{34.381} \\
        \bottomrule
        \end{tabular}
    \end{minipage}
    \vspace{-0.8cm}
\end{table}

%% file: tabs/coco_voc.tex
\begin{table}[t]
    \caption{\textbf{Unsupervised object segmentation on real-world datasets.} We compare SlotAdapt with state-of-the-art methods on VOC (left) and COCO (right). We present two versions of our method, with and without guidance loss.
    }
    \label{table:real-world-seg-quan}
    \centering
    \begin{subtable}
        \centering
        \small
        \setlength{\tabcolsep}{5pt}
        \begin{tabular}{lccc}
        \toprule
        \textbf{ VOC} & FG-ARI & mBO$^i$ & mBO$^c$ \\
        \midrule
        SA + DINO ViT & 12.3 & 24.6 & 24.9 \\
        SLATE + DINO ViT & 15.6 & 35.9 & 41.5 \\
        DINOSAUR & 23.2 & 43.6 & 50.8 \\
        LSD & 18.7 & 40.5 & 43.5 \\
        SlotDiffusion & 17.8 & 50.4 & \textbf{55.3} \\
        \midrule
        Ours & 28.8 & \textbf{51.6} & 52.0 \\ 
        Ours + Guidance & \textbf{29.6} & 51.5 & 51.9 \\
        \bottomrule
        \end{tabular}
    \end{subtable}
    \hfill
    \begin{subtable}
        \centering
        \small
        \setlength{\tabcolsep}{5pt}
        \begin{tabular}{lccc}
        \toprule
        \textbf{ COCO} & FG-ARI & mBO$^i$ & mBO$^c$ \\
        \midrule
        SA + DINO ViT & 21.4 & 17.2 & 19.2 \\
        SLATE + DINO ViT & 32.5 & 29.1 & 33.6 \\
        DINOSAUR & 34.3 & 32.3 & 38.8 \\
        LSD & 33.8 & 27.0 & 30.5 \\
        SlotDiffusion & 37.2 & 31.0 & 35.0 \\
        \midrule
        Ours & \textbf{42.3} & 31.5 & 34.8 \\
        Ours + Guidance & 41.4 & \textbf{35.1} & \textbf{39.2} \\
        \bottomrule
        \end{tabular}
    \end{subtable}
    \vspace{-0.4cm}
\end{table}

%% file: sections/conclusion.tex
\pagebreak
\vspace{-0.25cm}
\section{Conclusion}
\label{sec:conclusion} 
\vspace{-0.3cm}

We have targeted the object-centric learning problem, particularly on complex real-world images. To this end, we have presented a method that combines slot attention with pretrained diffusion models. Our architecture, SlotAdapt, has three main novelties; adapters in the diffusion model for slot conditioning, the use of a register token to represent background in images, and attention guidance to align slot attention with cross attention from the diffusion model. We have conducted extensive experiments, particularly on COCO dataset, to validate our approach. Our experiments show that  leveraging the generative capabilities of pretrained diffusion models is crucial for tackling the challenging tasks such as object discovery, unsupervised segmentation and compositional generation and editing on complex real-world images. Our method takes a step forward in achieving this goal, outperforming the state of the art methods on the challenging COCO dataset without relying on any external supervision in object discovery, segmentation and, particularly compositional generation and editing. We are the first to present successful compositional editing experiments on the COCO dataset. 

A limitation of our method from the perspective of compositional generation  is that the edited or reconstructed images, when conditioned on slots, may exhibit slight changes with respect to the source image, though the generations are mostly highly realistic and of very good quality. One remedy for this can be the incorporation of additional training objectives to enforce fidelity to the input. Another related issue is how to make use of this framework  for image editing in practice, which needs further adjustments to the architecture possibly to accept user prompts and associate them with slot representations. Lastly, our method still has issues for under- and over-segmentation, which may potentially be mitigated through slot merging and splitting. Additionally, our approach relies on pre-trained diffusion models that are predominantly trained on real-world data, which may limit their adaptability to synthetic domains. While the model performs well on natural images, its effectiveness may be reduced when handling synthetic datasets such as CLEVR and CLEVRTex~\cite{johnson2017clevr}.

%% file: supp/training_details.tex
\subsection{Training and Architectural Details}
\label{sec:train_details}
This section provides comprehensive information on the architectural components and training procedures of our model.

The importance of object-centric representations has been recognized across different domains, emerging as a future direction in works on temporal prediction~\cite{akan2021slamp} and autonomous driving~\cite{akan2022stretchbev}. Building on this insight, recent approaches have successfully applied diffusion models to achieve object-centric learning \cite{jiang2023lsd, wu2023slotdiffusion} and and our approach enhances this paradigm through more effective slot-based representations.

\subsubsection{Architectural Details}
Figure 1 in the main text illustrates our training pipeline. Below, we detail the key components:

\boldparagraph{Diffusion Model} We initialize a \unet denoiser and VAE from pretrained diffusion models \citep{rombach2022ldm}. Specifically:
\begin{itemize}
\item MOVi-E \citep{greff2022kubric} and COCO \cite{lin2014coco} datasets: Stable Diffusion v1.5
\item  VOC \citep{everingham2010pascalvoc}: Stable Diffusion v2.1
\end{itemize}
After initialization, we inject adapter layers following each downsampling and upsampling block in the \unet. Each adapter layer comprises a cross-attention mechanism and a feedforward network, both preceded by layer normalization.

\boldparagraph{VAE} We employ pretrained VAEs, maintaining consistency with the diffusion model versions: Stable Diffusion v1.5 VAE for MOVi-E and COCO; Stable Diffusion v2.1 VAE for VOC

\boldparagraph{Visual Encoder Backbone}
\begin{itemize}
\item MOVi-E: We utilize a custom CNN backbone encoder. It consists of 4 downsampling blocks, 1 middle block, and 4 upsampling blocks, with channel multipliers [1,1,2,4] and a base channel count of 128. Each block contains 2 residual layers and the overall network's output channels 128 channels.
\item COCO and VOC: We employ a frozen DINOv2 \citep{oquab2023dinov2} model with a ViT-B backbone (patch size 14).
\end{itemize}
The extracted feature maps maintain a consistent resolution of 32×32 across all datasets.

\boldparagraph{Slot Attention} Our implementation varies by dataset:
\begin{itemize}
\item MOVi-E: We adopt the LSD \citep{jiang2023lsd} configuration (slot size: 192, iterations: 3, slots: 24). We append four linear projectors to align slot dimensions with adapter attention layers.
\item COCO and VOC: We use a slot size of 768 and 7 slots, with similar output projector layers as in MOVi-E.
\end{itemize}
For all datasets, we use a linear layer to project either pooled visual backbone features or averaged slot vectors to a 768-dimensional space to match the text cross-attention dimensions in the diffusion model. We use slot averaging for MOVi-E and COCO datasets, and feature pooling for the VOC dataset, as they perform slightly better.

\boldparagraph{Training Details}
\begin{itemize}
\item Hardware: 2 NVIDIA A40 GPUs
\item Batch sizes: 32 (MOVi-E), 30 (VOC), 32 (COCO)
\item Training iterations: 200K (MOVi-E), 250K (COCO), 190K (VOC)
\item Optimization: AdamW optimizer, constant learning rate, FP16 precision
\item $\lambda$: In first 40K iterations, we use 0, from 40K-50K, it is increased from 0 to the $0.025$, then, we use $0.025$. Since the slots and adapter layers are initialized completely random, we opt to wait for 40K iterations so that the attention masks have meaningful connections.
\end{itemize}

\boldparagraph{Experimental Details}For our experiments, we first select the arhictectural details~(conditioning place and register token type) on MOVi-E dataset. Then, we migrate that architecture to the real-world setup. 

\boldparagraph{Dataset Details} For \textbf{MOVi-E}, we follow the train-test split in \cite{singh2022steve} and use $256\times256$ resolution for both diffusion model and slot attention. For \textbf{COCO}, we follow \cite{seitzer2023dinosaur} to train on the same training set with DINOSAUR, LSD and Slot Diffusion, which consists of 118,287 images for training and 5000 for validation. For \textbf{VOC}, we follow \cite{seitzer2023dinosaur} to train on the ``trainaug" set so that the training sets are the same as DINOSAUR, LSD and Slot Diffusion. The training set contains 10,582 images and validation set contains 1449 images. For COCO and VOC, we use random horizontal flip in training as data augmentation and we use $256\times256$ for diffusion model and $448\times448$ for slot attention.

%% file: supp/rebuttal_exps.tex
\subsection{Additional Experiments}
\label{sec:additional_exps}

\input{tabs/movie}

\vspace{-0.2cm}\boldparagraph{Synthetic Dataset Results} We evaluate the object discovery and segmentation performance of our method on MOVi-E in comparison to the baseline methods in \tabref{tab:movie} based on the FG-ARI, instance-level mBO and instance-level mIoU metrics. To measure the representational quality of the learned slots for downstream tasks, we implement a straightforward approach using a 2-layer MLP to predict discrete categories, object positions and 3D bounding boxes. We also display the resulting prediction accuracies in \tabref{tab:movie}. We observe significant improvements in almost all metrics, particularly with an approximate 10\% enhancement in object discovery and segmentation accuracy. 

\boldparagraph{Better Segmentation Model}In this part, we replace Slot Attention with BOQ-SA, which is an improved version of slot attention where the slot initialization is also optimized, ~\cite{jia2023boqsa}. The results~(in Table~\ref{tab:boqsa}) show that using BOQ-SA instead of Slot Attention results in modest improvements across several metrics.

\input{tabs/boqsa}

\boldparagraph{Slot Number} We investigate the impact of slot count by comparing our original configuration of 7 slots with an increased count of 80 slots, departing from the convention established in previous works \cite{seitzer2023dinosaur, wu2023slotdiffusion, jiang2023lsd} where slot count typically matches the maximum number of objects per scene in a given dataset. Our experiments revealed that dramatically increasing the slot count to 80 led to degraded performance, presented in Table~\ref{tab:slot_num}.

\input{tabs/slot_num}

\boldparagraph{Starting Guidance Loss from the start}We also experiment with starting of guidance loss, where the original idea was starting the guidance loss after some iterations so that the slots and the adapters have a knowledge of the objects. Below, we present results where the guidance loss is started in the beginning without waiting slots and adapters learn meaningful features. The results show that starting guidance loss after some iterations leads to better performance, presented in Table~\ref{tab:guidance_start}.

\input{tabs/guidance_start}

\boldparagraph{Additional Token for Slot Attention as a register token}We investigate an alternative approach to capturing global scene information in our architecture. Instead of using the average of all slot tokens as a global register token, we modified the slot attention module to include an additional dedicated slot specifically designed to capture global scene context. This dedicated slot was then directly used in the cross-attention layer. Our experimental results revealed interesting trade-offs: while the dedicated global slot approach improved foreground segmentation quality, it showed slightly decreased performance in semantic segmentation and generation quality. We hypothesize that this performance difference stems from the competitive nature of attention in the slot attention module---the dedicated global slot must compete with object slots for attention weights, potentially limiting its capacity to capture comprehensive scene information compared to our original approach of averaging all slot representations. This suggests that using the collective information from all slots through averaging provides a more robust global representation than a single learned global token, presented in Table~\ref{tab:additional_token}

\input{tabs/additional_token_exp}

\boldparagraph{Classifier-free Guidance}We conduct an ablation study for classifier-free guidance value, which simply increases the effect of the conditioning in diffusion models~\cite{ho2022cfg}. We find that CFG value of 1.3 results in the best results in terms both FID~\cite{heusel2017fid} and KID~\cite{binkowski2018kid}, presented in Table~\ref{tab:cfg_evals}

\input{tabs/cfg_evals}

%% file: tabs/movie.tex
\begin{table*}[t]
\begin{small}
    \caption{\textbf{Comparative evaluation on MOVi-E:}
    (Left) Segmentation results,
    (Right) Representation assessment: We evaluate slots through predictive probing. Spatial attributes (position, 3D bounding box) are assessed via MSE (mean squared error), while categorical predictions are assessed by classification accuracy.}
    \label{tab:movie}
    \centering
    \begin{adjustbox}{max width=\textwidth}
    \begin{tabular}{lcccc}
    \toprule
    \scriptsize{\textbf{Segmentation}}          & SLATE        & SLATE$^+$      & LSD & Ours   \\
    \midrule
    mBO ($\uparrow$)         & 30.17 & 22.17  & 38.96 & \textbf{43.38} \\
    mIoU ($\uparrow$)        & 28.59 & 20.63  & 37.64 & \textbf{41.85} \\
    FG-ARI ($\uparrow$)      & 46.06 & 45.25  & 52.17 & \textbf{56.45} \\
    \bottomrule
    \end{tabular}
        \begin{tabular}{lcccc}
    \toprule
    \scriptsize{\textbf{Representation}}           & SLATE        & SLATE$^+$      & LSD & Ours   \\
    \midrule
    Position ($\downarrow$)  & 2.09 & 2.15   & 1.85 & \textbf{1.77}  \\
    3D B-Box ($\downarrow$)   & 3.36 & 3.37  & \textbf{2.94} & 3.75    \\
    Category ($\uparrow$)    & 38.93 & 38.00 & 42.96 & \textbf{43.92} \\
    \bottomrule
    \end{tabular}
    \end{adjustbox}   
    \vspace{-0.5cm}
\end{small}
\end{table*}

%% file: tabs/boqsa.tex
\begin{table}[t]
   \caption{
   \textbf{Impact of a Better Segmentation Method:} We evaluate the effect of a better segmentation model on performance by replacing Slot Attention with BOQ-SA, an improved version. Results are presented on the COCO dataset.
   }
   \label{tab:boqsa}
   \centering
   \small
   \setlength{\tabcolsep}{5pt}
   \arrayrulecolor{black}
   
   \begin{tabular}{lcccccc}
       \toprule
       Method & 
       FG-ARI & 
       mBO$^i$ & 
       mIoU$^i$ & 
       mBO$^c$ & 
       mIoU$^c$ \\
       \midrule
       
       Slot Attention & 42.3 & 31.5 & 31.7 & 34.8 & 38.5 \\
       BOQ-SA & 42.2 & 31.2 & 32.551 & 35.266 & 37.82 \\
       
       \bottomrule
   \end{tabular}
\end{table}

%% file: tabs/slot_num.tex
\begin{table}[t]   
   \centering
   \small
   \setlength{\tabcolsep}{5pt}
   \arrayrulecolor{black}
   \caption{\textbf{Impact of Slot Count on Performance:} We analyze the effect of varying the number of slots on the COCO dataset.
   }
   \label{tab:slot_num}
   \begin{tabular}{lccccccc}
       \toprule
       \multirow{2}{*}{Method} & 
       \multirow{2}{*}{FG-ARI} & 
       \multicolumn{2}{c}{Instance} & 
       \multicolumn{2}{c}{Semantic} & 
       \multirow{2}{*}{FID} & 
       \multirow{2}{*}{KID} \\
       \cmidrule(lr){3-4} \cmidrule(lr){5-6}
       & & mBO & mIoU & mBO & mIoU & & \\
       \midrule
       
       80 slots & 18.1 & 23.2 & 26.4 & 28.6 & 32.9 & 114.236 & 64.610 \\
       7 slots & 41.4 & 35.1 & 36.1 & 39.2 & 41.4 & 10.857 & 0.388 \\
       
       \bottomrule
   \end{tabular}
\end{table}

%% file: tabs/guidance_start.tex
\begin{table}[ht]
   \caption{
    \textbf{Impact of Guidance Loss Timing:} We compare the effect of applying the guidance loss from the start of training versus introducing it after the model has learned initial representations. Results are presented on the COCO dataset.
    }
   \label{tab:guidance_start}
   
   \centering
   \small
   \setlength{\tabcolsep}{5pt}
   \arrayrulecolor{black}
   
   \begin{tabular}{lcccccc}
       \toprule
       \multirow{2}{*}{Method} & 
       \multirow{2}{*}{FG-ARI} & 
       \multicolumn{2}{c}{Instance} & 
       \multicolumn{2}{c}{Semantic} \\
       \cmidrule(lr){3-4} \cmidrule(lr){5-6}
       & & mBO & mIoU & mBO & mIoU \\
       \midrule
       
       Start from 0 & 37.85 & 32.65 & 33.99 & 36.059 & 39.254 \\
       Original (start after 40K) & 41.4 & 35.1 & 36.1 & 39.2 & 41.4 \\
       
       \bottomrule
   \end{tabular}
\end{table}

%% file: tabs/additional_token_exp.tex
\begin{table}[ht!]
   \caption{
    \textbf{Impact of Global Information Capturing Strategies:} We compare slot averaging and the use of an additional slot token for capturing global information. Results are presented on the COCO dataset.
    }
   \label{tab:additional_token}
   
   \centering
   \small
   \setlength{\tabcolsep}{4.5pt}
   \arrayrulecolor{black}
   
   \begin{tabular}{lccccccc}
       \toprule
       \multirow{2}{*}{Method} & 
       \multirow{2}{*}{FG-ARI} & 
       \multicolumn{2}{c}{Instance} & 
       \multicolumn{2}{c}{Semantic} & 
       \multirow{2}{*}{FID} & 
       \multirow{2}{*}{KID} \\
       \cmidrule(lr){3-4} \cmidrule(lr){5-6}
       & & mBO & mIoU & mBO & mIoU & & \\
       \midrule
       
       Additional Slot Token & 43.8 & 31.9 & 32.4 & 35.5 & 37.3 & 11.212 & 0.431 \\
       Slot Average Token & 42.3 & 31.5 & 34.8 & 34.8 & 38.5 & 10.857 & 0.388 \\
       
       \bottomrule
   \end{tabular}
\end{table}

%% file: tabs/cfg_evals.tex
\begin{table}[t]
    \caption{\textbf{Impact of CFG Value on Generation Quality:} We evaluate the effect of different CFG values on generation quality using the COCO dataset.
    }
    \label{tab:cfg_evals}
    \centering
    \small
    \setlength{\tabcolsep}{5pt}
    
    \arrayrulecolor{black}
    
    \begin{tabular}{lcc}
        \toprule
        CFG Value & FID & KID$\times$1000 \\
        \midrule
        
        7.5 & 21.350 & 6.271 \\
        5.0 & 17.558 & 4.236 \\
        2.5 & 13.734 & 2.151 \\
        2.0 & 12.427 & 1.424 \\
        1.5 & 11.041 & 0.590 \\
        1.4 & 10.880 & 0.459 \\
        1.3 & \textbf{10.857} & \textbf{0.388} \\
        1.2 & 11.057 & 0.492 \\
        
        \bottomrule
    \end{tabular}
\end{table}

%% file: supp/qual_examples.tex
\subsection{Additional Qualitative Examples}
\label{sec:qual}

In this section, we present additional qualitative examples to further illustrate the capabilities of our proposed SlotAdapt model. These results complement the main paper by providing a more comprehensive view of our model's performance across various tasks and datasets.

\begin{figure*}[!ht]
\centering
\includegraphics[width=1.\textwidth]{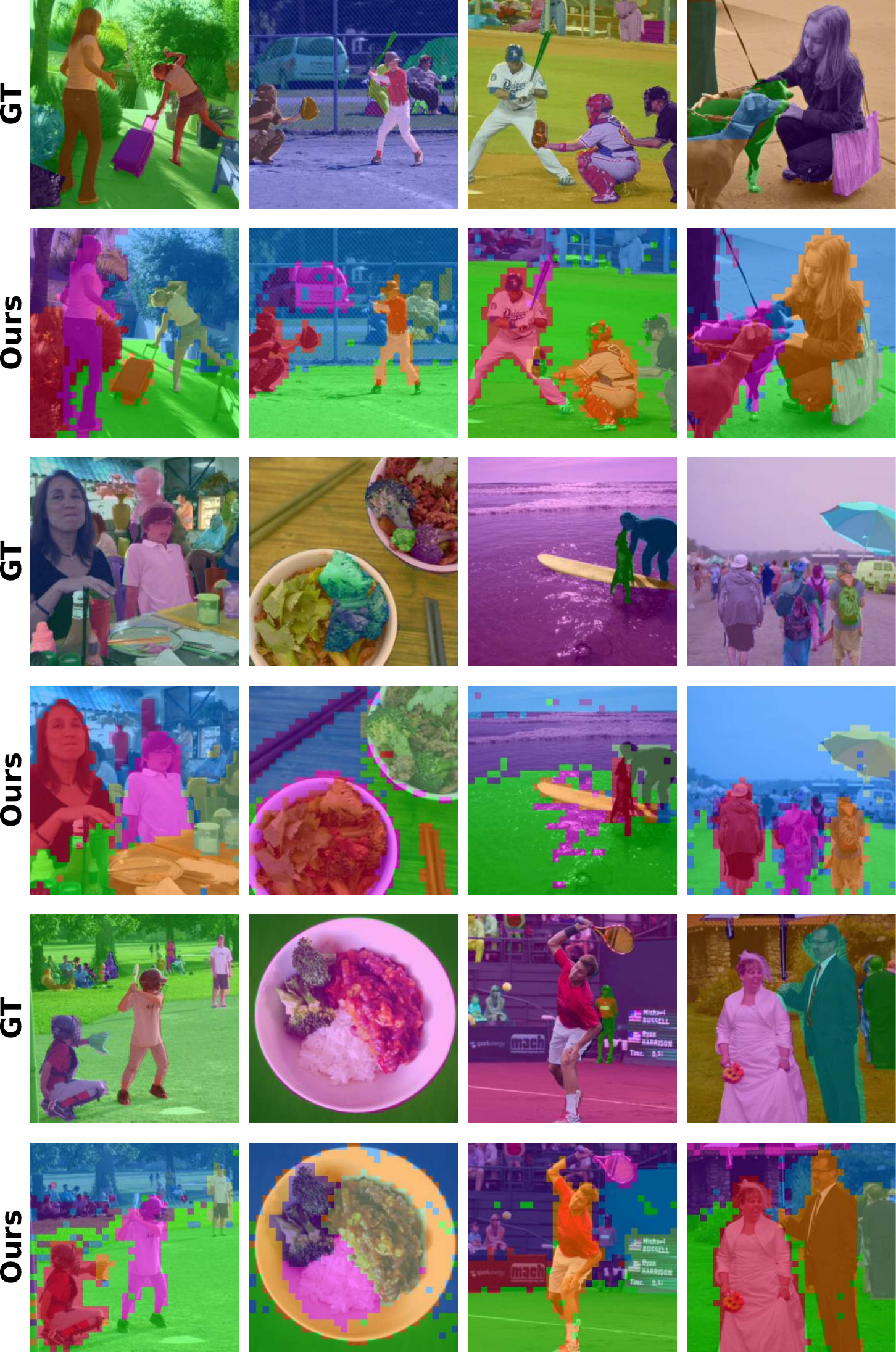}
\vskip -0.1in
\caption{\textbf{Unsupervised Object Segmentation.} We show visualizations of predicted segments on COCO.}
\vspace{-0.4cm}
\label{fig:supp_coco_seg}
\end{figure*}

\begin{figure*}[!ht]
\centering
\includegraphics[width=1.\textwidth]{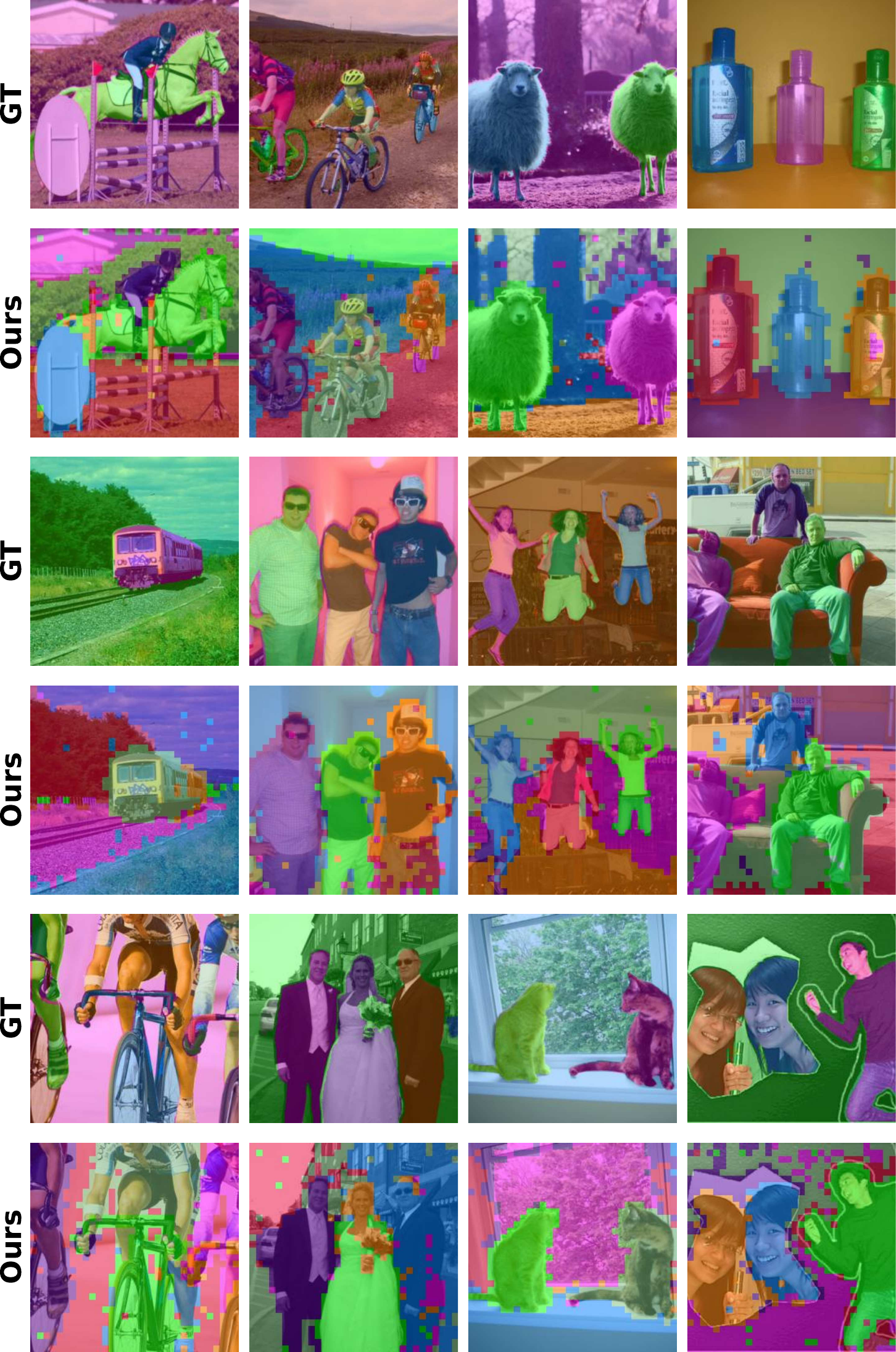}
\vskip -0.1in
\caption{\textbf{ Unsupervised Object Segmentation.} We show visualizations of predicted segments on  VOC.}
\vspace{-0.4cm}
\label{fig:supp_voc_seg}
\end{figure*}

\begin{figure*}[!ht]
\centering
\includegraphics[width=1.\textwidth]{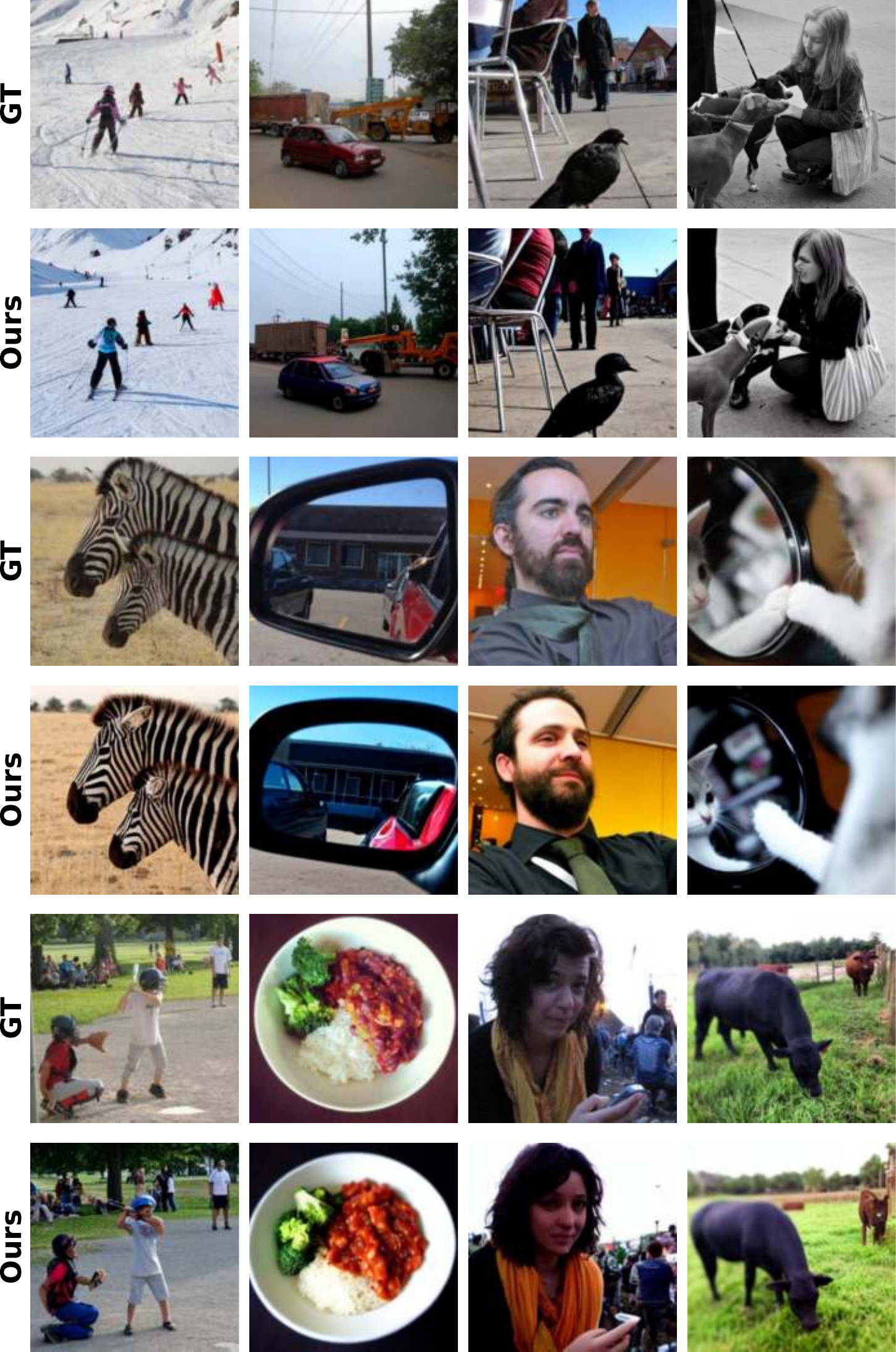}
\vskip -0.1in
\caption{\textbf{Generation Results.}
We show sample images from COCO, reconstructed by SlotAdapt conditioned on slots.}
\vspace{-0.4cm}
\label{fig:supp_coco_gen}
\end{figure*}

\begin{figure*}[!ht]
\centering
\includegraphics[width=1.\textwidth]{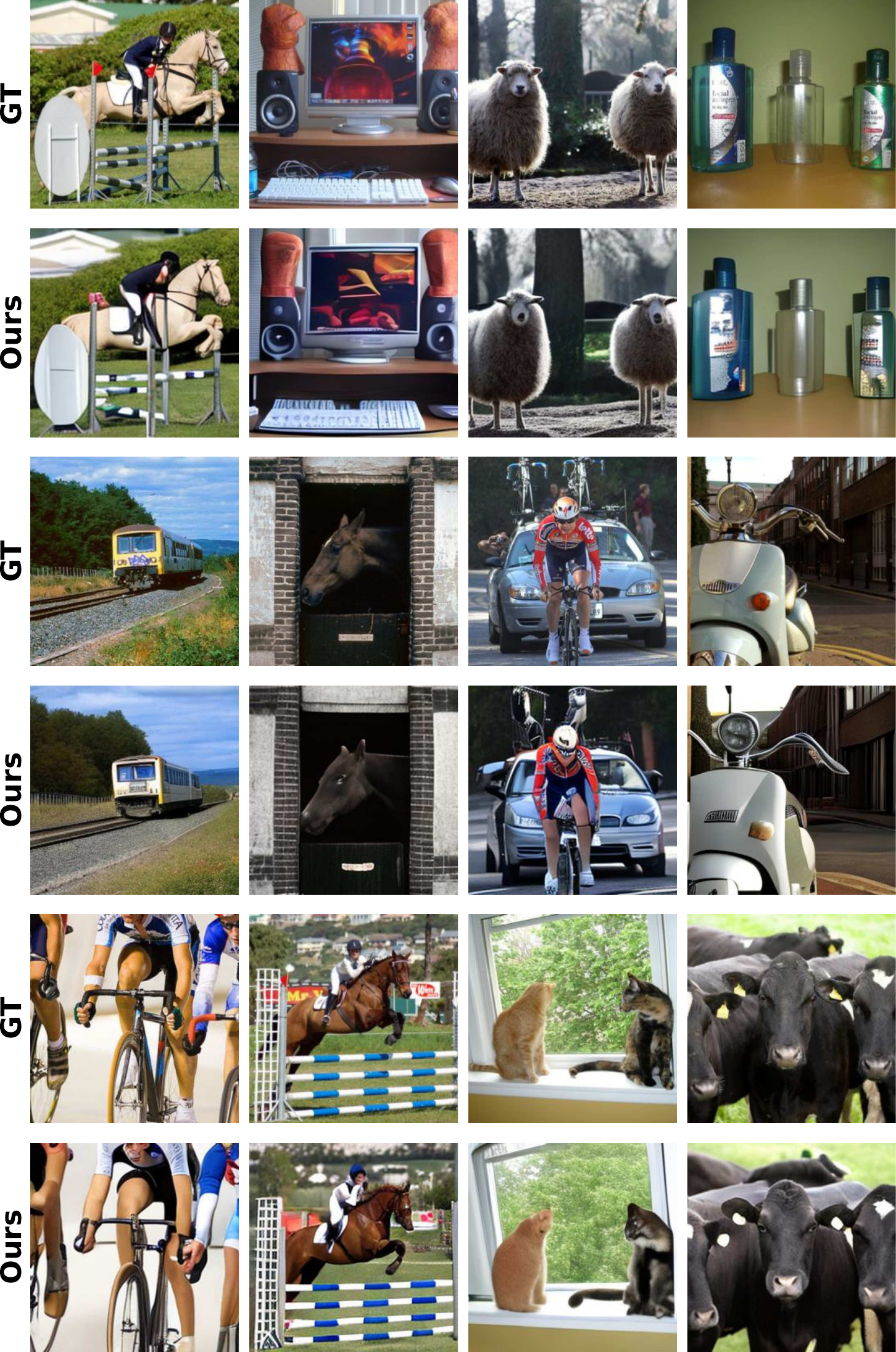}
\vskip -0.1in
\caption{\textbf{Generation Results.} We show sample images from VOC, reconstructed by SlotAdapt conditioned on slots.}
\vspace{-0.4cm}
\label{fig:supp_voc_gen}
\end{figure*}

\begin{figure*}[!ht]
\centering
\includegraphics[width=1.\textwidth]{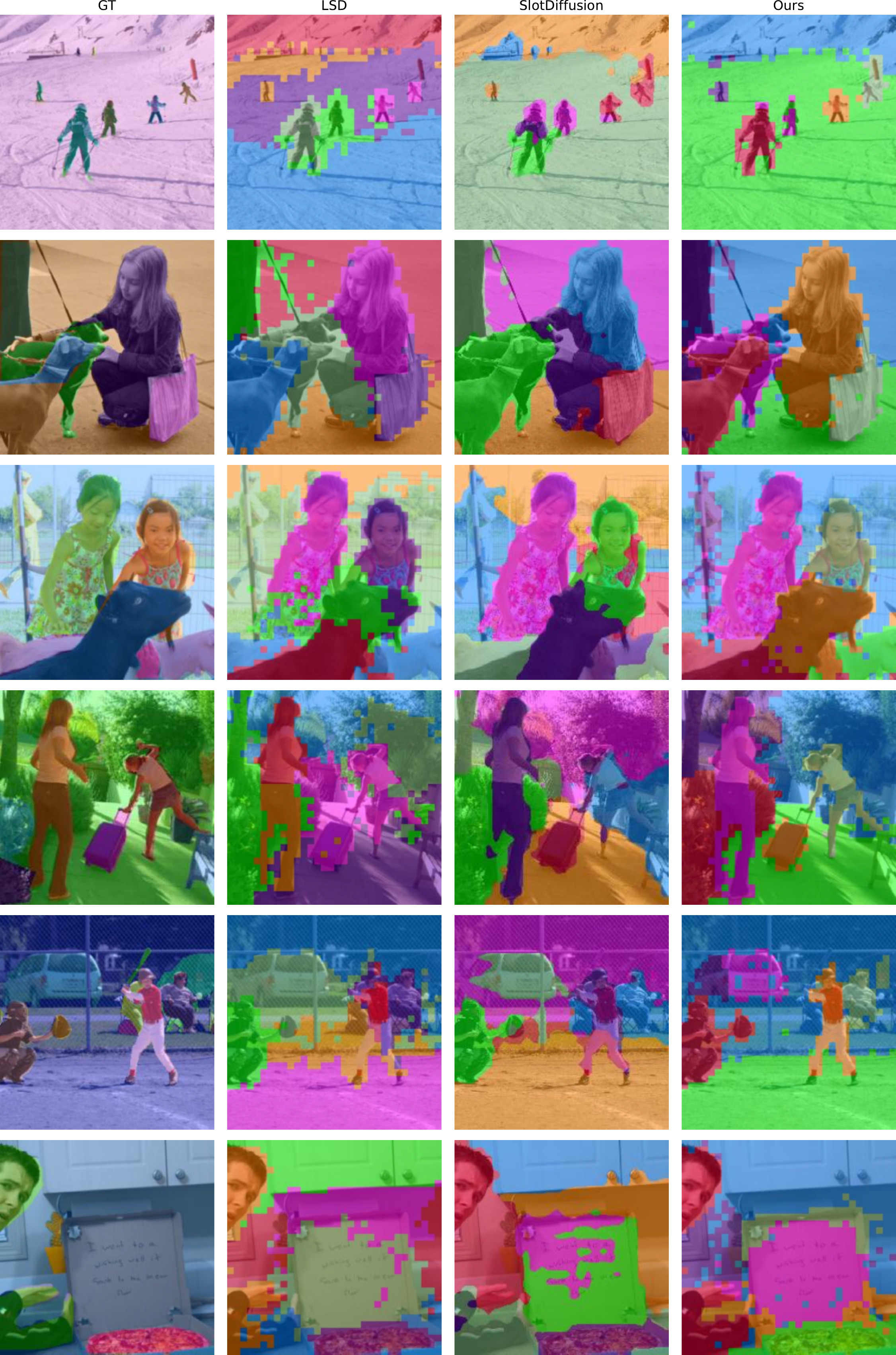}
\vskip -0.1in
\caption{\textbf{Segmentation comparisons with other methods} We show visualizations of predicted segments on COCO dataset. Compared to other models, our model tends to produce more coherent masks with fewer fragmented segments.}
\vspace{-0.4cm}
\label{fig:supp_coco_attn_compare}
\end{figure*}

\begin{figure*}[!ht]
\centering
\includegraphics[width=1.\textwidth]{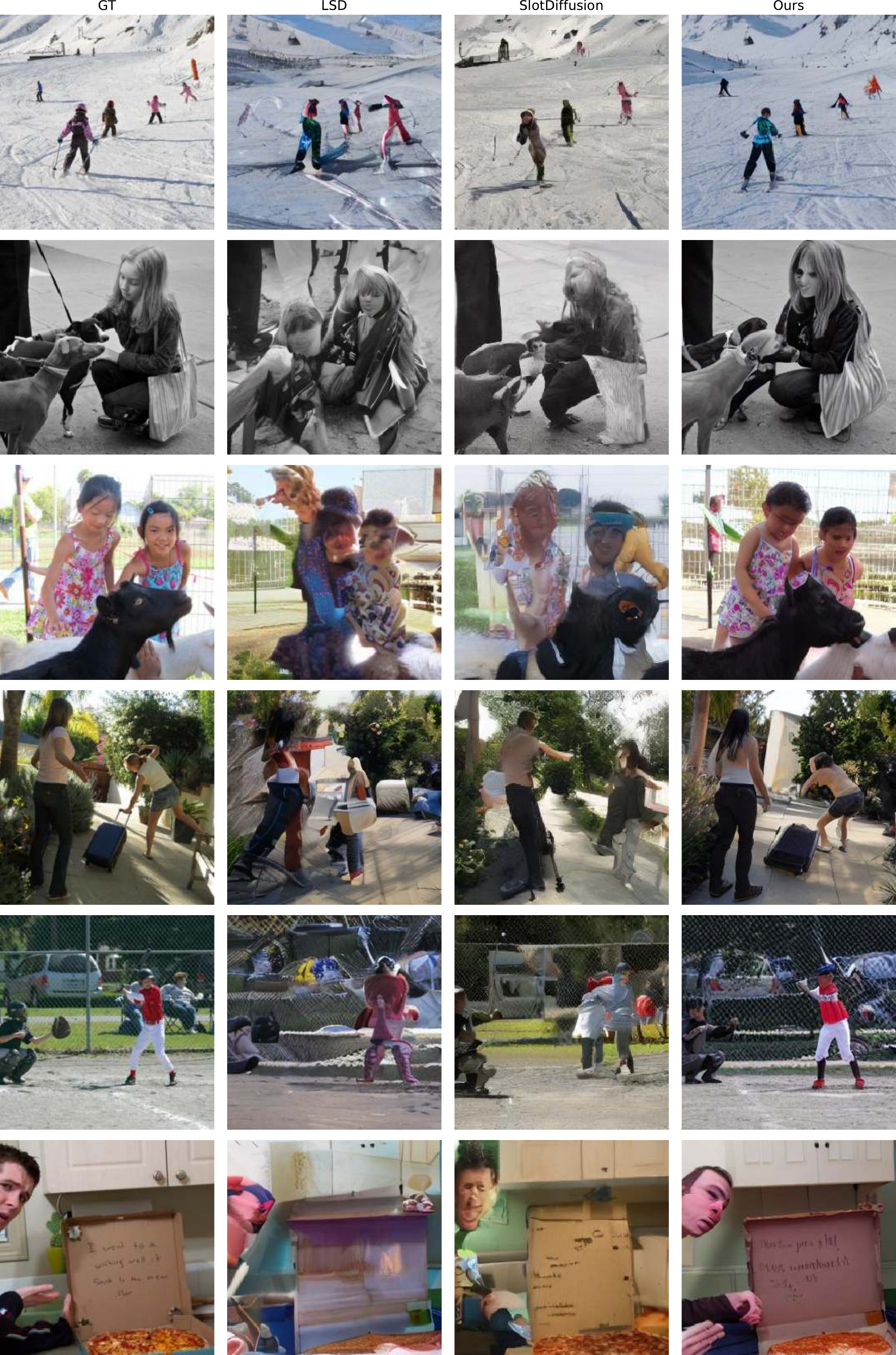}
\vskip -0.1in
\caption{\textbf{Generation comparisons with other methods.} We show visualizations of generated images on COCO dataset. Compared to other models, our model can generate better reconstructions.}
\vspace{-0.4cm}
\label{fig:supp_coco_gen_compare}
\end{figure*}

\begin{figure*}[!ht]
\centering
\includegraphics[width=1.\textwidth]{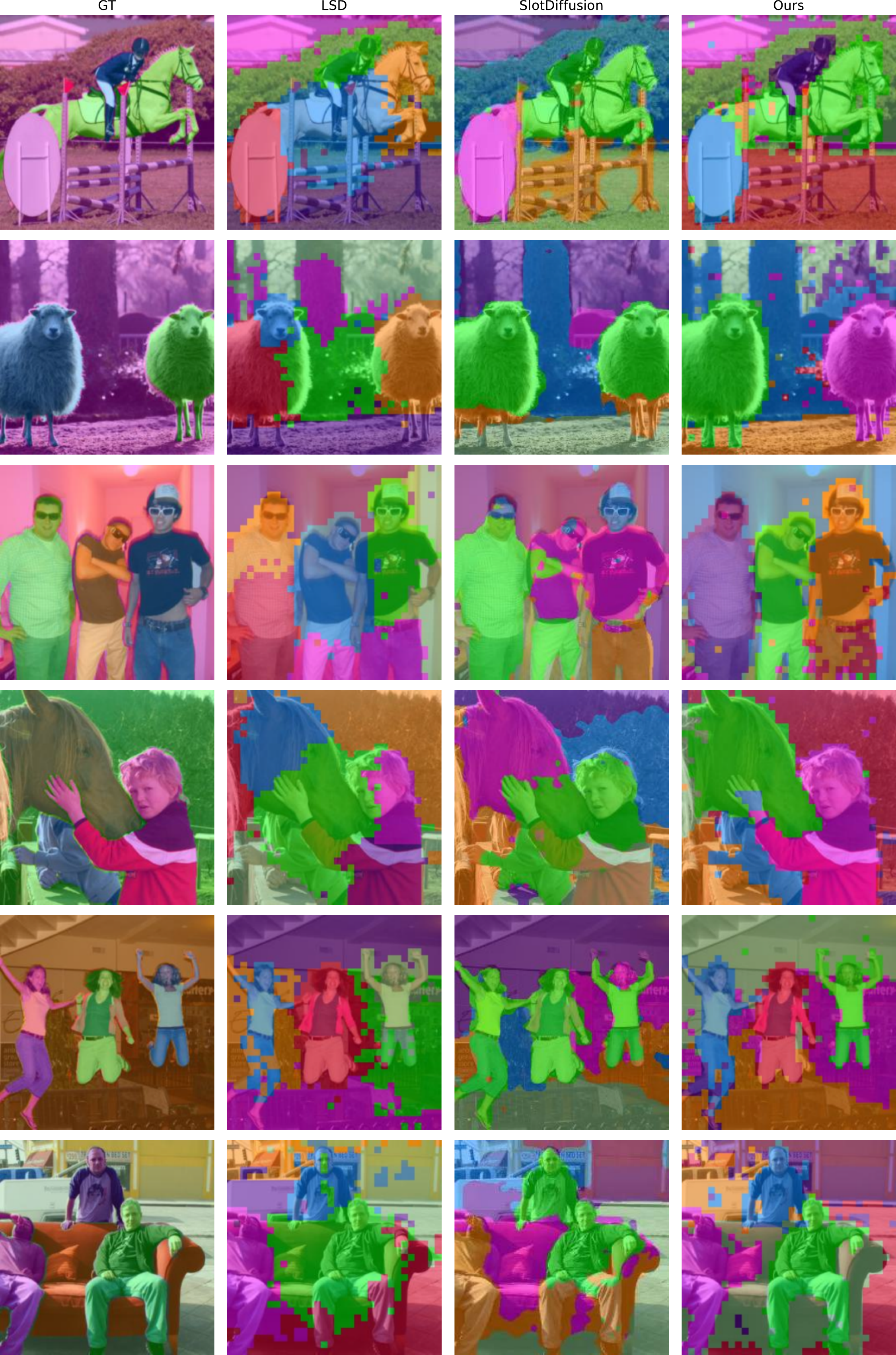}
\vskip -0.1in
\caption{\textbf{Segmentation comparisons with other methods.} We show visualizations of predicted segments on  VOC dataset. Compared to other models, our model tends to produce more coherent masks with fewer fragmented segments.}
\vspace{-0.4cm}
\label{fig:supp_voc_attn_compare}
\end{figure*}

\begin{figure*}[!ht]
\centering
\includegraphics[width=1.\textwidth]{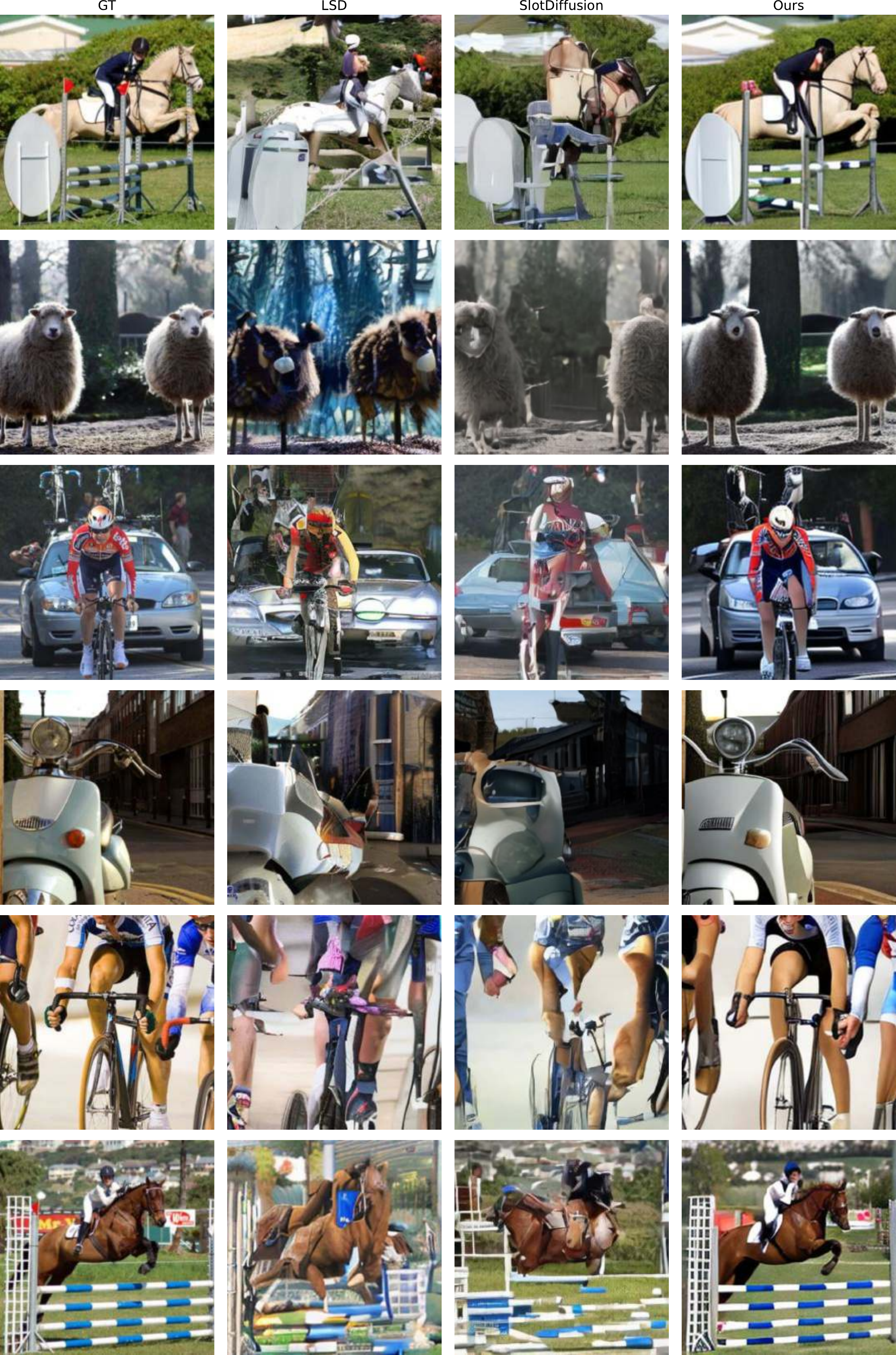}
\vskip -0.1in
\caption{\textbf{Generation comparisons with other methods.} We show visualizations of generated images on  VOC dataset. Compared to other models, our model can generate better reconstructions.}
\vspace{-0.4cm}
\label{fig:supp_voc_gen_compare}
\end{figure*}

\begin{figure*}[!ht]
\centering
\includegraphics[width=0.97\textwidth]{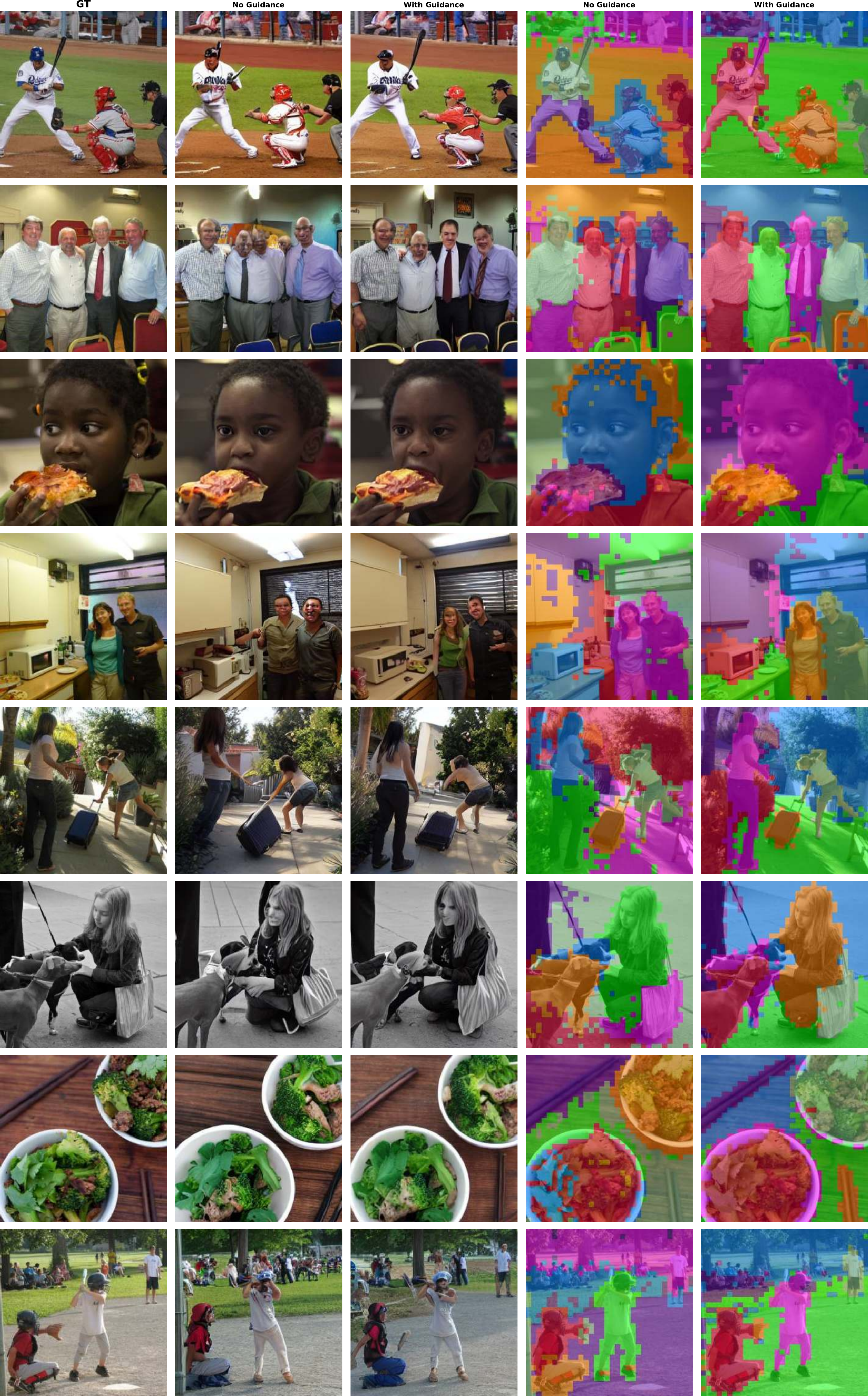}
\vskip -0.1in
\caption{\textbf{Qualitative comparison: with vs. without guidance.} We visualize generated images and predicted segments on COCO dataset. Guidance helps to generate better aligned objects and to mitigate  the ``part-whole" hierarchy problem in object segmentation task.}
\vspace{-0.4cm}
\label{fig:supp_guidance_compare}
\end{figure*}

\begin{figure*}[!ht]
\centering
\includegraphics[width=1.\textwidth]{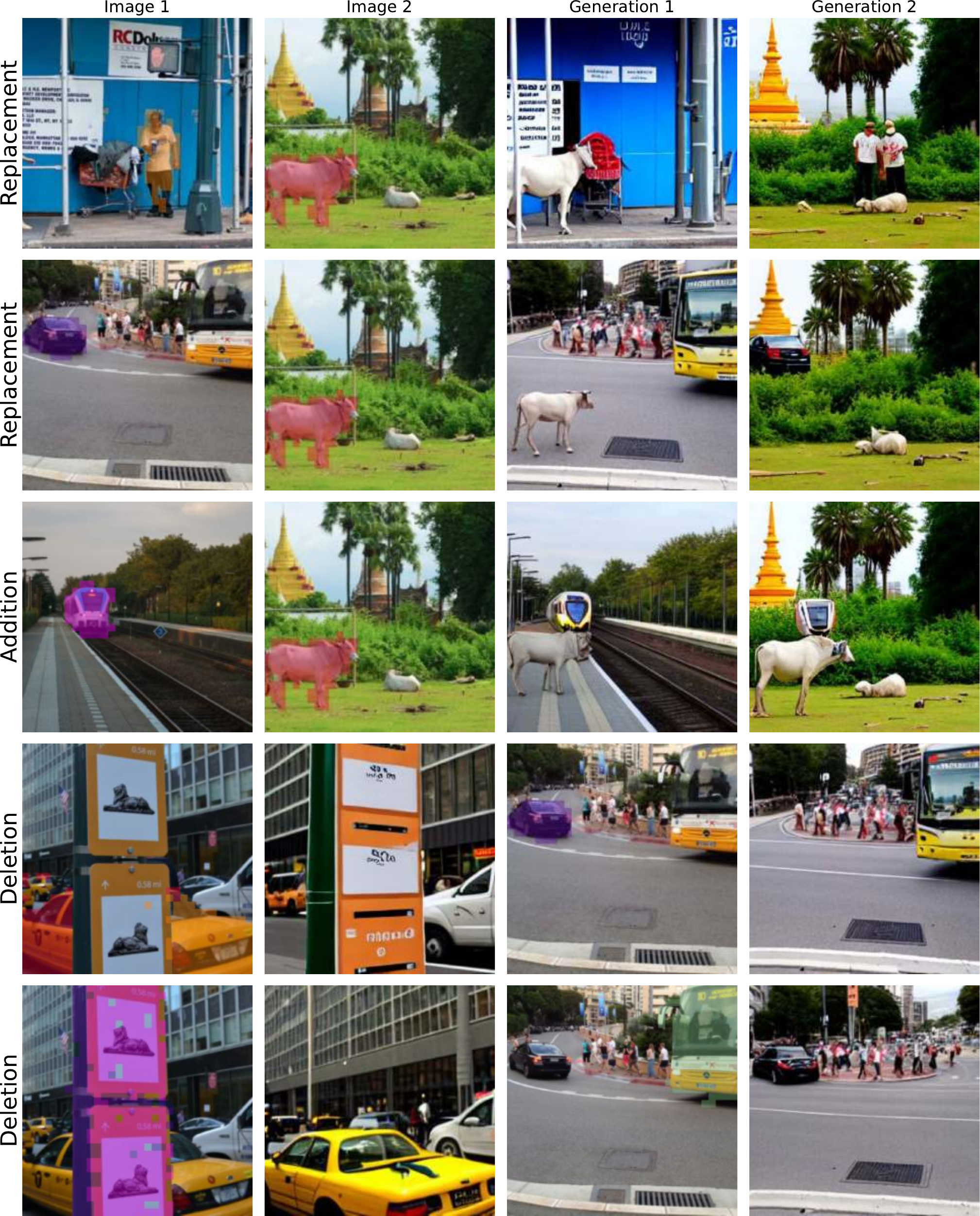}
\vskip -0.1in
\caption{\textbf{Compositional Generations and Editing.} We show visualizations of generated images when the slots are manipulated. In the first two rows, we exchange the highlighted slots. In the middle row, we simply add the highlighted slot to the other image. In the last two rows, we delete the highlighted slots to manipulate the image. For all scenarios, our model can successfully generate realistic images in complex scenarios. This shows our model's ability for compositional generation and editing.}
\vspace{-0.4cm}
\label{fig:supp_comp_gen}
\end{figure*}

%% file: supp/glass_comp.tex
\subsection{Comparison with GLASS and SPOT}
\label{sec:glass}

This section presents a comprehensive comparison between our method and the concurrent work GLASS \citep{singh2024guidedsa} and SPOT \citep{Kakogeorgiou2024SPOT}, highlighting both quantitative and qualitative differences.

\boldparagraph{Methodological Distinctions} GLASS leverages extra information, such as class labels or image captions, to enhance its performance. While this approach yields certain advantages, it introduces limitations. Primarily, GLASS struggles to differentiate between instances of the same class due to its reliance on semantic masks as pseudo ground truth. Additionally, the need for additional information restricts the  applicability of GLASS in fully unsupervised scenarios.
In contrast, our method operates without any external supervision, successfully segments individual instances even within the same class, and captures nuanced scene information without relying on pre-defined semantic categories. We should also note that GLASS does not present any compositional editing results.

SPOT operates in a fully unsupervised manner by introducing (i) an attention-based self-training mechanism that distills improved slot-based attention masks from the decoder to the encoder, and (ii) a patch-order permutation strategy for autoregressive transformers to better utilize slot representations during reconstruction. Although SPOT achieves strong object segmentation performance, particularly on complex real-world images, it still faces challenges in accurately differentiating fine-grained object instances within the same class. We should also note that SPOT does not present any compositional editing or reconstruction results.

\begin{figure*}[!ht]
\centering
\includegraphics[width=.8\textwidth]{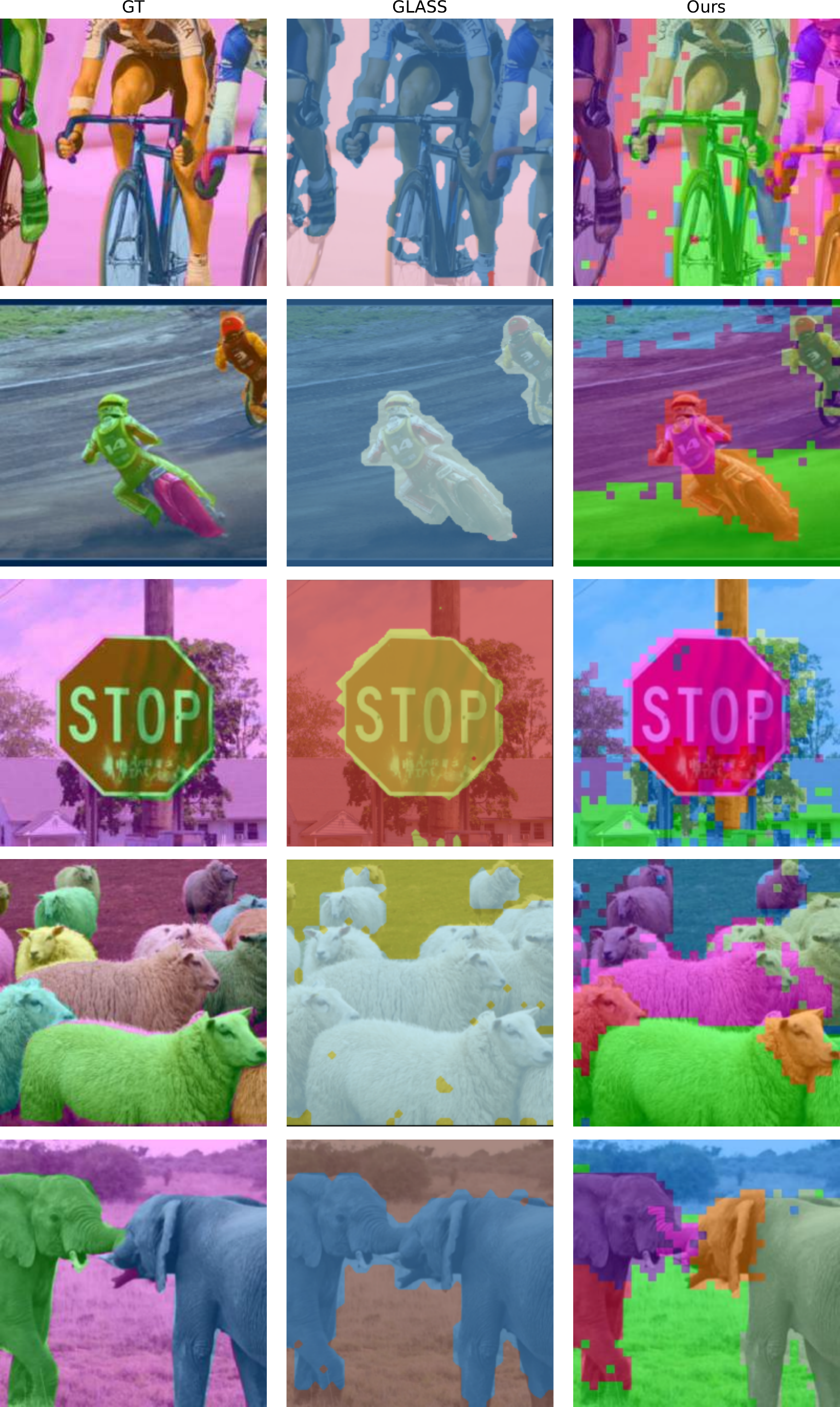}
\vskip -0.1in
\caption{\textbf{Unsupervised Object Segmentation.} We show visualizations of predicted segments for SlotAdapt vs. GLASS on real world datasets (VOC and COCO).}
\vspace{-0.4cm}
\label{fig:supp_glass_compare}
\end{figure*}

\Figref{fig:supp_glass_compare} provides visual comparisons on both COCO and  VOC datasets. Our model demonstrates superior performance in instance differentiation, accurately segmenting multiple instances of the same class (as shown in rows 1, 2, and 4). Furthermore, it excels in scene understanding, capturing meaningful elements such as trees and house roofs (row 3) without explicit labeling.

\begin{figure*}[!ht]
\centering
\includegraphics[width=.8\textwidth]{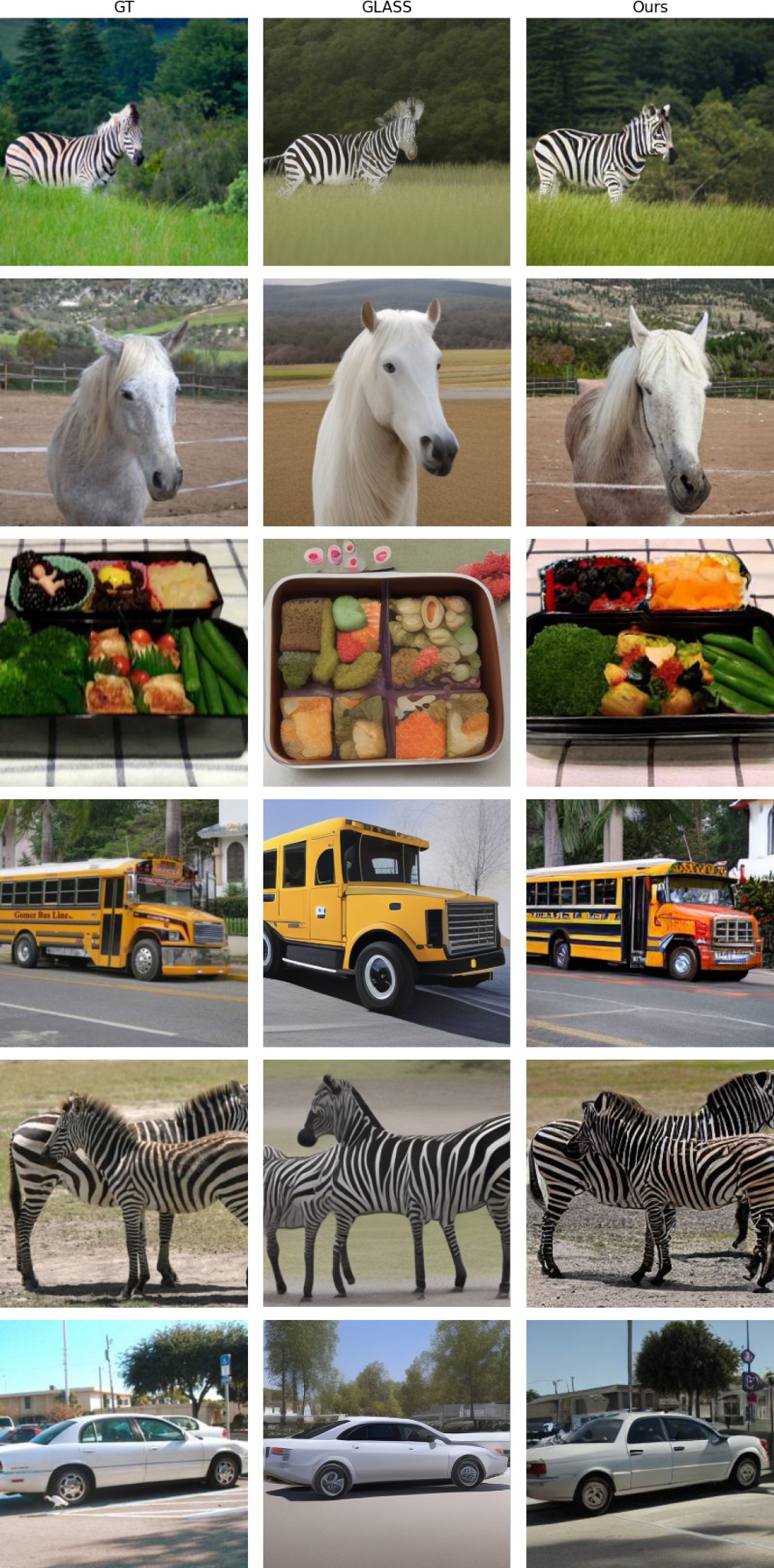}
\vskip -0.1in
\caption{\textbf{Generation results.} We show visualizations of generated images by SlotAdapt vs. GLASS on COCO and VOC.}
\vspace{-0.4cm}
\label{fig:supp_glass_compare_gen}
\end{figure*}

Table \ref{tab:glass-compare} presents a quantitative comparison of our method with GLASS and SPOT. In terms of semantic overlap, GLASS excels, which is attributable to its use of semantic masks for supervision. However, on complex datasets like COCO, which features multiple instances per image, SlotAdapt achieves comparable performance without any additional supervision. SPOT, while fully unsupervised, focuses on enhancing slot representations through self-training and patch-order permutation strategies. Although it performs well in capturing global scene structures, particularly on COCO, it lags behind our method in fine-grained instance differentiation, as reflected in the FG-ARI scores.

These results underscore the robustness and versatility of our unsupervised approach, particularly in handling complex, multi-instance scenarios. Our method demonstrates that high-quality object segmentation can be achieved without relying on external supervision, offering a more flexible and generalizable solution for diverse image understanding tasks.

\input{tabs/supp_glass}

%% file: tabs/supp_glass.tex
\begin{table}[!ht]
    \vspace{-5mm}
    \caption{
    Unsupervised object segmentation comparisons with the concurrent work GLASS and SPOT on VOC (left) and COCO (right). We would like to point out that both GLASS and GLASS$^\dagger$ use extra supervision such as class labels or image caption. SPOT is a fully unsupervised method.
    \label{tab:glass-compare}
    }
    \centering
    \begin{subtable}
        \centering
        \small
        \setlength{\tabcolsep}{5pt}
        \begin{tabular}{lccc}
        \toprule
        \textbf{PASCAL VOC} & FG-ARI & mBO$^i$ & mBO$^c$ \\
        \midrule
        Ours & 28.8 & 51.6 & 52.0 \\ 
        Ours + Guidance & \textbf{29.6} & 51.5 & 51.9 \\
        GLASS$^\dagger$ & \textemdash & \textbf{60.4} & \textbf{68.4} \\
        GLASS & \textemdash & 58.1 & 36.1 \\
        SPOT w/o ENS & 19.7 & 48.1 & 55.3 \\
        SPOT w/ ENS & 19.9 & 48.3 & 55.6 \\
        \bottomrule
        \end{tabular}
    \end{subtable}
    \hfill
    \begin{subtable}
        \centering
        \small
        \setlength{\tabcolsep}{5pt}
        \begin{tabular}{lccc}
        \toprule
        \textbf{MS COCO} & FG-ARI & mBO$^i$ & mBO$^c$ \\
        \midrule
        Ours & \textbf{42.3} & 31.5 & 34.8 \\
        Ours + Guidance & 41.4 & 35.1 & 39.2 \\
        GLASS$^\dagger$ & \textemdash & 34.3 & 45.2 \\
        GLASS & \textemdash & \textbf{35.3} & \textbf{46.3} \\
        SPOT w/o ENS & 37.8 & 34.7 & 44.3 \\
        SPOT w/ ENS & 37.8 & 35.0 & 44.7 \\
        \bottomrule
        \end{tabular}
    \end{subtable}
\end{table}